\documentclass{article} 
\usepackage{iclr2025_conference,times}

\usepackage{amsmath,amsfonts,bm}

\def\eqref#1{equation~\ref{#1}}

\def\1{\bm{1}}

\def\rmW{{\mathbf{W}}}
\def\rmX{{\mathbf{X}}}

\def\ermA{{\textnormal{A}}}

\def\ermK{{\textnormal{K}}}

\def\ermP{{\textnormal{P}}}
\def\ermQ{{\textnormal{Q}}}
\def\ermR{{\textnormal{R}}}
\def\ermS{{\textnormal{S}}}

\def\va{{\bm{a}}}

\def\vl{{\bm{l}}}

\def\vr{{\bm{r}}}

\def\vx{{\bm{x}}}

\DeclareMathAlphabet{\mathsfit}{\encodingdefault}{\sfdefault}{m}{sl}
\SetMathAlphabet{\mathsfit}{bold}{\encodingdefault}{\sfdefault}{bx}{n}

\usepackage{hyperref}
\usepackage{url}
\usepackage[pdftex]{graphicx}
\usepackage{subcaption}
\usepackage{algorithm}
\usepackage{algorithmic}
\usepackage{fontawesome}
\usepackage{soul}

\usepackage[capitalize]{cleveref}

\usepackage{xspace}
\newcommand{\ouralgo}{\mbox{\emph{A-MoD}}\xspace}
\newcommand{\ouralgoheading}{\mbox{\textup{\emph{A-MoD}}}\xspace}  

\usepackage{tikz}
\usetikzlibrary{positioning,arrows,fit,shapes.geometric}
\pgfdeclarelayer{background}
\pgfdeclarelayer{foreground}
\pgfsetlayers{background,main,foreground}

\title{Attention Is All You Need For Mixture-of-Depths Routing}

\author{
{\bf Advait Gadhikar\textsuperscript{\rm 1, \rm 2 \faEnvelope}, 
Souptik Majumdar\textsuperscript{\rm 1, \rm 3 \faEnvelope}, 
Niclas Popp\textsuperscript{\rm 1}}
Piyapat Saranrittichai\textsuperscript{\rm 1}, \\
{\bf 
Martin Rapp\textsuperscript{\rm 1},
Lukas Schott\textsuperscript{\rm 1}}, \\
\textsuperscript{\rm 1}Bosch Center for Artificial Intelligence, Renningen, Germany \\
\textsuperscript{\rm 2}CISPA Helmholtz Center for Information Security, Saarbrücken, Germany \\
\textsuperscript{\rm 3}University of Stuttgart, Stuttgart, Germany \\
\textsuperscript{\rm 3}University of Tübingen, Tübingen, Germany \\
\texttt{\{advait.gadhikar@cispa.de, st184540@stud.uni-stuttgart.de\}}
}

\iclrfinalcopy 
\begin{document}

\maketitle

\begin{abstract}
Advancements in deep learning are driven by training models with increasingly larger numbers of parameters, which in turn heightens the computational demands.
To address this issue, Mixture-of-Depths (MoD) models have been proposed to dynamically assign computations only to the most relevant parts of the inputs, thereby enabling the deployment of large-parameter models with high efficiency during inference and training.
These MoD models utilize a routing mechanism to determine which tokens should be processed by a layer, or skipped.
However, conventional MoD models employ additional network layers specifically for the routing which are difficult to train, and add complexity and deployment overhead to the model.
In this paper, we introduce a novel attention-based routing mechanism \ouralgo that leverages the existing attention map of the preceding layer for routing decisions within the current layer.
Compared to standard routing, \ouralgo allows for more efficient training as it introduces no additional trainable parameters and can be easily adapted from pre-trained transformer models.
Furthermore, it can increase the performance of the MoD model.
For instance, we observe up to $2\%$ higher accuracy on ImageNet compared to standard routing and isoFLOP ViT baselines. Furthermore, \ouralgo  improves the MoD training convergence, leading to up to $2\times$ faster transfer learning. 

\end{abstract}

\section{Introduction}
\begin{figure}
    \centering    \includegraphics[width=0.6\linewidth]{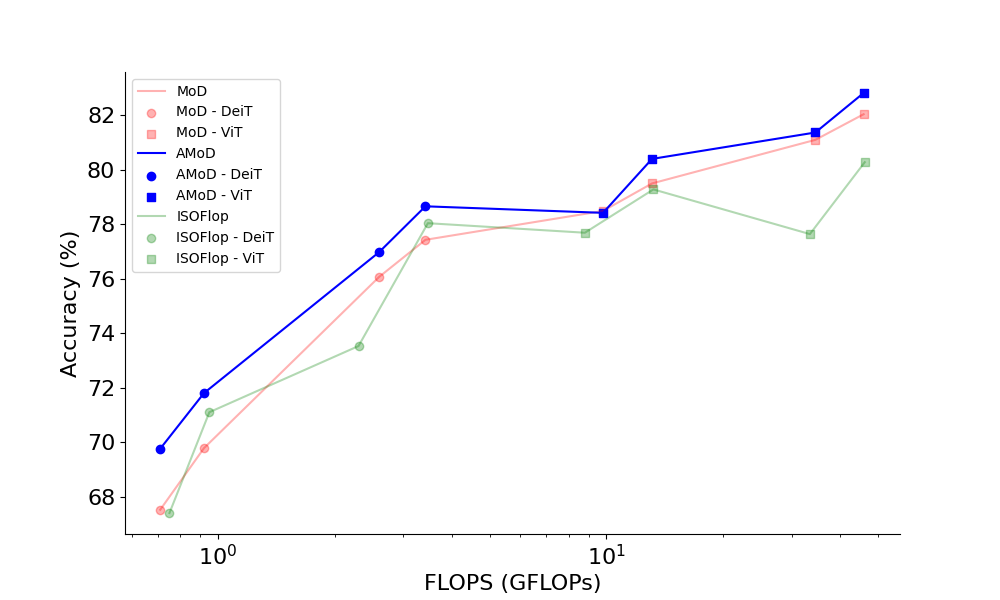}
    \caption{Accuracy vs FLOPs Pareto-curve for \ouralgo in comparison with MoD and ISOFlop models on ImageNet-1k.}
    \label{fig:pareto}
\end{figure}

Increasing the model size has enabled transformer-based deep learning models to achieve state-of-the-art performance across various domains, including computer vision \citep{dosovitskiy2021an} and natural language processing \citep{hoffmann2022training, kaplan2020scaling} – even unlocking emergent capabilities \citep{wei2022emergent}. 
However, the computational costs of these large models present significant challenges \citep{thompson2020computational}. 
Therefore, reaching a Pareto-optimal model to maximize both efficiency and performance is crucial (see \cref{fig:pareto}).

\citet{jacobs1991adaptive} originally introduced conditional computation via mixture of experts, laying the foundations to increase model sizes while maintaining FLOPs, by dynamically activating only a subset of the model parameters, termed experts, conditioned on the input. This principle allowed scaling towards outrageously large networks \citep{shazeer2016outrageously} and is leveraged at the forefront of current Large Language Models (LLMs) \citep{jiang2024mixtral}.

Compared to standard deep learning models \cite{dosovitskiy2021an,wang2024yolov10,he2016deep}, dynamic models have received less research attention and are often not yet competitive on the Pareto front of performance and runtime on standard GPU architectures.  
Here, we focus on further advancing the field of dynamic compute. 

\begin{figure}
    \centering
    \input{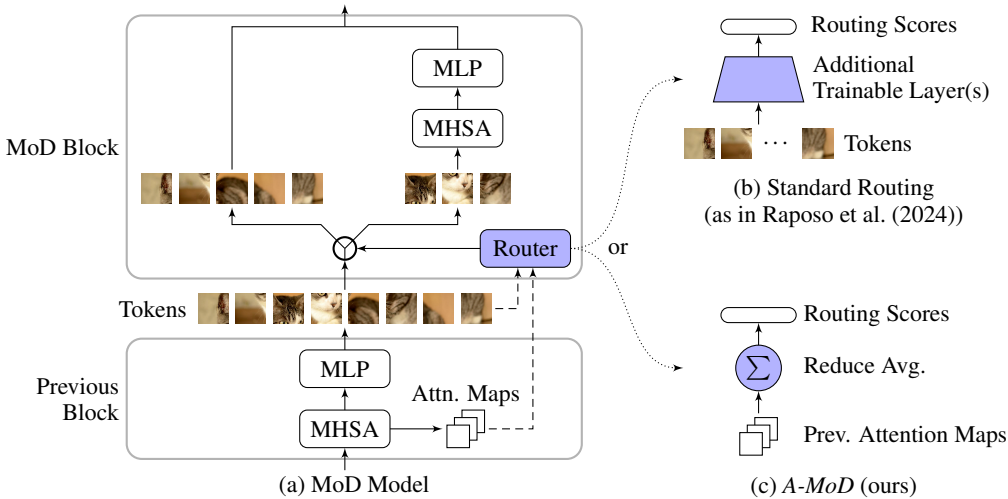}
    \caption{MoD model (a) with standard routing (b) vs.\ our \emph{A-MoD} attention routing~(c).}
    \label{fig:overview}
\end{figure}

Recently, \citet{raposo2024mixture} introduced Mixture-of-Depths (MoD) as a variant of mixture of experts. 
In MoD models, the computational costs are dynamically reduced by processing only a subset of tokens in a layer while the remaining tokens skip the layer (see \cref{fig:overview}a). Compared to baselines with equivalent FLOPs, MoDs can perform favorably on language tasks.
A crucial component of MoD is its router, which receives tokens as inputs and, given a user-defined capacity, determines which tokens should enter or skip a layer.
The router usually consists of a linear layer that is jointly trained along with the model (\cref{fig:overview}b).  

The routing mechanisms heavily influence the model performance for several reasons. First, routing introduces noise into the training process, as the routing is a discrete decision and is often performed at multiple layers and per token. 
Second, routers depend on additional layers, and hence, need to be trained from scratch when adapting a vanilla pretrained model to an MoD model.
Lastly, the router adds a small computational overhead to the sparsified model. 

Hence, in this paper, we ask and address the question:
\textit{Can we improve the routing mechanism in MoD models based on information that is already available within the model, instead of using additional trainable parameters within the router?}
We find the answer to our question in the attention mechanism of commonly used transformer architectures \citep{vaswani2017attention,dosovitskiy2021an}.

We assume that the attention maps can be used to estimate the importance of a token, by averaging its interaction with other tokens.
Based on that, we propose to aggregate the information in the attention maps and use it as an importance measure for token routing in MoD.
We call our method attention routing for MoD: \ouralgo (\cref{fig:overview}c). 
We find that \ouralgo can outperform standard routing in MoD networks across a range of model sizes and tasks consistently (as shown in \cref{fig:pareto}).
Not only is our \ouralgo parameter-free, but it can also be applied to adapt off-the-shelf pretrained transformer models to MoDs with almost no additional training. 
We further validate our method empirically and show that routing scores computed by \ouralgo are better correlated with token importance estimates compared to routing scores from standard routers.

This paper presents a significant advancement in the application of MoD to the visual domain.Our primary contributions are:
\begin{itemize}
    \item We find that MoD is not only viable but also advantageous for visual tasks, providing empirical evidence that it can outperform traditional models in terms of both FLOPs and performance.
    \item We introduce \ouralgo, a parameter-free routing method for MoDs based on the attention maps to compute token importance.
    \item We demonstrate that \ouralgo outperforms a standard router across a range of datasets for MoDs on finetuning and transfer learning. In the case of transfer learning, \ouralgo exhibits faster convergence of MoD models transferred from a dense pretrained model.
    \item Compared to standard MoD, \ouralgo consistently selects important tokens, and routing decisions correlate with leave-one-out token importance that is estimated by removing tokens.
\end{itemize}


\section{Related Work}
\paragraph{Attention Maps}
The attention mechanism \citep{bahdanau2014neural} enables models to learn long-range dependencies within sequences with a constant number of operations. 
Transformers \citep{vaswani2017attention} leverage the attention mechanism in the language domain and have become a de facto standard model.
\citet{dosovitskiy2021an} further adapted transformers to the vision domain by treating image patches as tokens, introducing the vision transformer (ViT). 
For images, attention maps have been shown to focus on key areas such as objects in the image \citep{carion2020end,jetley2018learn}. 
In this paper, we utilize this property for effective routing of tokens within neural networks. Furthermore, we use the Data-efficient image Transformers DeiT-T and DeiT-S \citep{deit} instead of vanilla ViT-T and ViT-S models, as small ViTs do not generalize well when trained on smaller datasets \citep{dosovitskiy2021an}. 

\paragraph{Mixture of Experts and Mixture-of-Depths}
Since their introduction over three decades ago \citep{jacobs1991adaptive,og_moe}, Mixture of Experts (MoE) have been applied to various model types. \citet{shazeer2016outrageously} introduced MoEs to scale transformer architectures \citep{ludziejewskiscaling}.
Subsequently, MoEs have achieved extensive empirical success across vision and language tasks \citep{puigcerversparse, jain2024mixture, fedus2022review, riquelme2021scaling}. One of the main challenges when training MoE networks is training instability \citep{zoph2202st,fedus2022switch}. \citet{raposo2024mixture} recently introduced the Mixture-of-Depths (MoD) architecture, where each transformer block processes only a subset of tokens, achieving a favorable compute-performance trade-off compared to large transformer models. \citet{liangevit} also induce sparsity by fusing tokens together without entirely skipping tokens. In their current form, both MoEs and MoDs use dedicated routing networks that decide which components of the overall network process which tokens. The difference between MoEs and MoDs is that an MoE model comprises several distinct experts that independently process the tokens. In contrast, an MoD model only chooses between  two experts per layer one of which is the layer itself and the other an identity function, see \cref{fig:overview}.


\paragraph{Routing Methods}
Routing mechanisms are required for most conditional computation blocks \citep{cai2024survey}. In MoE models for transformers, the purpose of the router is to match tokens to experts such that performance is maximized. In case of models with a single expert such as Switch Transformers \citep{fedus2022switch} or MoD, routers decide whether a token will benefit from processing by the expert or will be skipped. Various methods \citep{liurouters} have been proposed such as learned routers \citep{shazeer2016outrageously} with token choice or expert choice routing \citep{zhou2022mixture}, solving a linear program to match tokens to experts \citep{lewis2021base}, hashing inputs to match experts \citep{roller2021hash} or  using reinforcement learning to make routing decisions \citep{clark2022unified, bengio2015conditional, bengio2013estimating}. Explicitly learning the routers is the current state-of-the-art that outperforms other methods in most cases \citep{dikkala2023benefits}. However, this approach mainly proves effective with a larger number of routing parameters and is prone to training instabilities \citep{ramachandran2018diversity}. Thus, training routers that consistently lead to strong performance remains an open problem.

Our work focuses on improving the MoD architecture. We propose a novel routing mechanism, based on attention maps,  thereby eliminating the need for a standard router. The tokens are routed in a parameter-free manner without any extra computational overhead.

\section{Method}

In this section, we explain the Mixture-of-Depths (MoD) architecture and introduce our attention-based MoD routing algorithm, \ouralgo, that can be employed to improve its routing.


\subsection{Mixture-of-Depths}

Our work focuses on Vision Transformers \citep{dosovitskiy2021an,deit}. Here, given an input in terms of tokens $\rmX$, the output predictions are calculated by a model $f(\rmX; \Theta)$ consisting of $L$ Transformer blocks parameterized by a set of learnable weights $\Theta$. Each transformer block includes a Multi-Head Self-Attention (MHSA) with $H$ heads, followed by a two-layers fully-connected network with GeLU activations (MLP).

In MoD, \citet{raposo2024mixture} introduce a variation of transformer-based architectures with the assumption that individual tokens require varying amounts of compute within a model. 
In particular, MoD layers only process a subset of selected important tokens, while the remaining tokens skip the layer. Empirically, this procedure can improve the performance over a vanilla ViT with a comparable compute budget. 

Whether or not tokens skip a layer is determined by token importance scores estimated by a routing algorithm. Conventionally, standard routing computes these importance scores with additional layers (see Section \ref{sec:method_standard_routing}). In contrast, our \ouralgo computes the scores directly from the attention maps of previous layers without the need of additional parameters (see Section \ref{sec:method_attention_routing}).

\subsection{Standard routing}
\label{sec:method_standard_routing}

Considering a single MoD layer, the standard approach to compute the importance scores of input tokens requires an additional router network, as shown in Figure \ref{fig:overview}b. Typically, a router is a linear layer that projects a token vector to a scalar representing its importance score (as introduced by \citet{raposo2024mixture}). Formally, we consider the $l$-th transformer layer $f_l(\rmX^{l-1}; \theta_l)$ parameterized by a set of parameters $\theta_l$ with an input $\rmX^{l-1} = \left[ \vx_1^{l-1}; \vx_2^{l-1}; \ldots; \vx_N^{l-1} \right] \in \mathbb{R}^{N \times d}$ representing a token sequence of length $N$. Now, we can estimate token importance scores as:
\begin{align}
\vr_i = (\rmX^{l-1}\rmW_r^l)_i,
\end{align}
where $\rmW_r^l \in \mathbb{R}^{d \times 1}$ is the parameter of the additional linear routing network. These tokens will be skipped or processed based on their scores as per the equation below:

\begin{align}
\label{eq:method_mod_standard}
    \vx_i^{l} =  \begin{cases} 
    r_i f_l\left(\rmX^{l-1}\right)_i + \vx_i^{l-1} & \text{if }\;\; \vr_i \geq P_{\beta}(\ermR^l) \\
     \vx_i^{l-1} & \text{else} 
    \end{cases}
\end{align}
Here, $P_{\beta}(\ermR^l)$ denotes the $\beta$-th percentile of all token importance scores $\ermR^l$. $\beta$ can be defined in terms of the capacity $C$ as $\beta := 1 - \frac{C}{N}$, where $C \in \left(0, 1\right)$ is the capacity for the MoD layer.
To learn the token importance scores during backpropagation, the output of the transformer layer is multiplied by the importance scores $r_i$, such that it can receive a non-zero gradient.

\subsection{Attention routing}
\label{sec:method_attention_routing}

In contrast to standard routing, we propose \ouralgo, a method to compute routing scores based on attention without additional trainable parameters. 
\ouralgo leverages the attention map of the previous layer to determine the routing scores for the current MoD layer, as shown in Figure \ref{fig:overview}c. The attention map $\ermA^{l-1}_h \in \mathbb{R}^{N \times N}$ of the $h$-th head from the previous layer can be computed as follows \citet{vaswani2017attention}:
\begin{align}
\label{eq:method_attention}
\ermA^{l-1}_h = \text{softmax}\left( \frac{(Q^{l-1}_h)(K^{l-1}_h)^T}{\sqrt{d}}\right),
\end{align}
where $Q^{l-1}_h  \in \mathbb{R}^{N \times d}$ and $K^{l-1}_h  \in \mathbb{R}^{N \times d}$ are query and key matrices computed from the previous layer respectively, and $d$ is the embedding dimension of query and key.

Following Equation \ref{eq:method_attention}, each element $\va^{l-1}_{h, ji}$ of $\ermA^{l-1}_h$ indicates how much information from the $i$-th token is considered when computing the $j$-th output. 
Aggregating $\va^{l-1}_{h, ji}$ across all rows yields a measure for the relevance of the $i$-th token with respect to all other tokens.
Therefore, in \ouralgo, we propose to compute a token importance score by averaging the corresponding attention values across all rows and attention heads as:
\begin{align}
\vr_i = \frac{1}{HN}\sum_{h=1}^H\sum_{j=1}^N\va^{l-1}_{h, ji}.
\end{align}

Based on the score computation above, the output from the $l$-th layer can then be calculated as:
\begin{align}
\label{eq:method_mod_attention}
    \vx_i^{l} =  \begin{cases} 
    f_l\left(\rmX^{l-1}\right)_i + \vx_i^{l-1} & \text{if }\;\; \vr_i \geq P_{\beta}(\ermR^l) \\
     \vx_i^{l-1} & \text{else} 
    \end{cases}
\end{align}
We note that, for \ouralgo, we do not multiply the token scores $\vr_i$ by the output, as the attention maps are already learnable in the previous layer. 
This preserves the original token output, promoting faster training when adapting from a vanilla pretrained checkpoint. We also tried a variation with multiplying $\vr_i$, but this did not lead to performance improvements and, therefore, was removed in the favor of simplicity. 
For standard routing in Equation \ref{eq:method_mod_standard}, this multiplication term is required to properly calculate the gradient of the router parameters. 
In contrast, \ouralgo removes the parameters of the router and thereby enables easier post-hoc adaptation of MoDs and eliminates training instabilities of routing scores.


\section{Experiments}

\subsection{Training setup and overview}

In our experiments, we systematically evaluate \ouralgo and empirically demonstrate its benefits over standard routing for MoDs. 
We perform evaluations across a range of model architectures and multiple image classification tasks.
In each experiment, we train a MoD, adapted from a vanilla pretrained transformer model.
We conduct experiments on both finetuning the adapted MoD model on the same dataset used for training the vanilla pretrained transformer model and transfer learning on different datasets.

\paragraph{Training setup}
We evaluate \ouralgo across four vision transformer architectures of varying sizes: DeiT-Tiny, DeiT-Small \citep{deit}, ViT-Base and ViT-Large \citep{dosovitskiy2021an}. 
Each MoD architecture is adapted from a vanilla pretrained checkpoint on ImageNet-1k \citep{imagenet}. 
Starting from this checkpoint, we train the MoD models with $50\%$ and $12.5\%$ capacity as described in \cref{eq:method_mod_attention}, i.e., $50\%$ and $12.5\%$ tokens are processed in each MoD layer, respectively.
Following \citet{raposo2024mixture}, we alternate between MoD layers and dense layers in our MoD architecture i.e. every second layer is an MoD layer.
We also analyze the effect of placing MoD layers only in the later layers of the model in \cref{sec:mod-layer-position}.

\paragraph{Finetuning}
We finetune the MoD models on ImageNet-1k.
For each case, we compare our \ouralgo to standard routing.
We also compare both MoD variants to an isoFLOP vanilla vision transformer.
This isoFLOP model is obtained by appropriately reducing the number of layers of the original model to match the number of FLOPs of its MoD counterpart.
Only reducing the layers still allows the isoFLOP model to benefit from the weights of the pretrained checkpoint.
Each model is trained with the AdamW optimizer \citep{adamw} for $100$ epochs using a batch size of $128$ and a learning rate of $1e-5$ with a linear warmup followed by cosine annealing.
We identify this learning rate schedule after performing a sweep as shown in \cref{fig:lr-deit-s,fig:lr-deit-t,fig:lr-sweep-vit} in the Appendix. 

\paragraph{Transfer learning}
To further investigate the benefits of \ouralgo, we perform transfer learning for image classification on the smaller Stanford Cars \citep{cars}, Oxford Pets \citep{pets} and Flowers102 \citep{flowers} datasets.
Here, each model is trained with SGD for $200$ epochs, a batch size of $64$ and learning rate $0.01$ with cosine annealing.

\paragraph{Token importance}
Finally, we conduct a comparison between the routing scores computed by standard routing and \ouralgo, with a reference score that measures the importance of each token.
This analysis enables us to further distinguish the benefits of \ouralgo.
We use a leave-one-out method \citep{hastie2009elements} to estimate the token importance. In particular, we measure the \emph{change in the loss of the model if a certain token is removed at an MoD layer}.
This allows us to assign a reference importance score to each token in each MoD layer, for every input image. We then correlate this with the  our routing weights for each MoD layer and token, both, for our attention-based routing and standard routing.  
Overall, not only does \ouralgo choose visually relevant tokens, but the routing scores also correlate strongly with the leave-one-out token importance.

\subsection{\ouralgoheading improves performance for finetuning}

For finetining, we train each MoD model on ImageNet. Across all our considered  vision transformer models (ranging from 5M to 300M parameters), \ouralgo mostly outperforms standard routing.
Results for MoDs with $50\%$ and $12.5\%$  capacity are presented in \cref{tab:accuracy-in1k}.
Through the training curves presented In \cref{fig:convergence-in1k} for $50\%$ capacity and \cref{fig:convergence-in1k-app} for $12.5\%$ capacity in the Appendix we highlight that \ouralgo converges faster. 

\begin{table}
    \centering
    \caption{\ouralgo \textbf{mostly outperforms MoD with standard routing and the isoFLOP baseline on ImageNet}, both for $50\%$ and $12.5\%$ capacity.}
    \begin{tabular}{l l c c c c}
        \toprule
       \multirow{2}{*}{\textbf{Model}} & \multirow{2}{*}{\textbf{Configuration}} & \multicolumn{2}{c}{\textbf{C}$=12.5\%$} &  \multicolumn{2}{c}{\textbf{C}$=50\%$} \\
               & & \textbf{FLOPs (G)} & \textbf{Accuracy (\%)} & \textbf{FLOPs (G)} & \textbf{Accuracy (\%)}   \\

        \midrule
        \multirow{3}{*}{DeiT-Tiny} 
        & isoFLOP            & 0.75 & 67.4  & 0.95 & 71.1 \\
        & MoD   & 0.71 & 67.52 & 0.92 & 69.78 \\
        & \ouralgo  & 0.71 & \textbf{69.76} & 0.92 & \textbf{71.8} \\
        \midrule
        \multirow{3}{*}{DeiT-Small} 
        & isoFLOP            & 2.3 & 73.53  & 3.47 & 78.04 \\
        & MoD   & 2.6 & 76.07  & 3.42 & 77.43 \\
        & \ouralgo  & 2.6 & \textbf{76.98}  & 3.42 & \textbf{78.66} \\
        \midrule
        \multirow{3}{*}{ViT-Base}  
        & isoFLOP            & 8.8 & 77.69  & 13.21 & 79.28 \\
        & MoD   & 9.8 & \textbf{78.49}  & 13.1 & 79.5 \\
        & \ouralgo  & 9.8 & 78.42  & 13.1 & \textbf{80.4} \\
        \midrule
        \multirow{3}{*}{ViT-Large} 
        & isoFLOP            & 33.4 & 77.64    & 46.24 & 80.28 \\
        & MoD   & 34.5 & 81.1  & 45.92 & 82.04 \\
        & \ouralgo  & 34.5 & \textbf{81.37} & 45.92 & \textbf{82.82} \\
        \bottomrule
    \end{tabular}
    \label{tab:accuracy-in1k}
\end{table}

For the DeiT-Tiny model with $50\%$ capacity (see \cref{fig:convergence-deit-t}), \ouralgo outperforms MoD by more than $2\%$ and by $1\%$ on the other larger models.
Similarly, for $12.5\%$ capacity, \ouralgo outperforms standard routing on both DeiT-Tiny and Small and is on par for the larger variants.
While \ouralgo is marginally worse for the ViT-Base model for $12.5\%$ capacity, it requires fewer epochs to converge as shown in the convergence plots in \cref{fig:convergence-vit-b-125} (in the Appendix) and already achieves this peak at the $20$-th epoch. Overall, \cref{tab:accuracy-in1k} along with the training curves in \cref{fig:convergence-in1k} confirm that \ouralgo can outperform MoDs with standard routing as well as isoFLOP baselines.
Specifically, \ouralgo has larger performance improvements for the smaller DeiT-Tiny and DeiT-Small and enables faster convergence across all models.

\paragraph{Adapting from pretrained checkpoints}
As described in \cref{eq:method_mod_attention}, \ouralgo can compute routing scores solely based on the attention maps and it does not multiply the output of each MoD block with the routing score, thus mostly conserving the token output.
Both properties allow \ouralgo finetuned from a pretrained checkpoint with attention routing to converge with minimal training.
\cref{fig:convergence-in1k} illustrates that \ouralgo enables much faster convergence, greatly reducing the required training time compared to standard routing.
In some cases, \ouralgo can achieve reasonable accuracy without any training.
This is exemplified in \cref{fig:convergence-vit-b}, where \ouralgo achieves $78\%$ accuracy without any training.
All accuracies of MoDs adapted from a pretrained checkpoint without training are reported in \cref{tab:accuracy-adapt} in the Appendix and highlight that \ouralgo always starts from a higher accuracy than standard routing. 
This is possible as the model estimates the least important tokens using the already learned attention maps, such that final accuracy is minimally affected as further substantiated in \cref{sec:token-importance}.
In contrast, standard routing multiplies layer outputs by the routing scores and needs to learn routing from scratch, as it is based on additional layers.
These factors result in slower convergence.

\textbf{Multiplying routing scores to output in \ouralgo} In \cref{fig:attn-multiply} in the Appendix, we compare \ouralgo with a modification that multiplies the output of the MoD block with the attention routing score to verify if \ouralgo benefits from an additional learned gradient like standard routing, i.e., using \cref{eq:method_mod_standard} instead of \cref{eq:method_mod_attention}.
However, multiplying the routing scores to the output for \ouralgo, instead, worsens accuracy of the adapted MoD model without any training and slightly slows down convergence.

\textbf{Learning rate stability analysis} To investigate the stability of our training with respect to the learning rate for \ouralgo and MoD, we perform a sweep over various learning rates and track the performance. We find that for all tested individual learning rates,  \ouralgo outperforms MoD, see \cref{fig:lr-sweep-vit}. 

\begin{figure}
    \centering
    \subfigure[DeiT-T]{
        \includegraphics[width=0.22\linewidth]{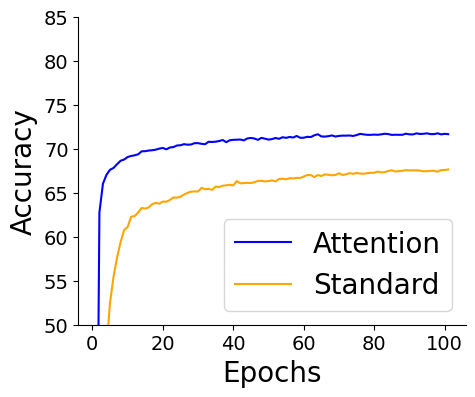}
        \label{fig:convergence-deit-t}
    }
    \subfigure[DeiT-S]{
        \includegraphics[width=0.22\linewidth]{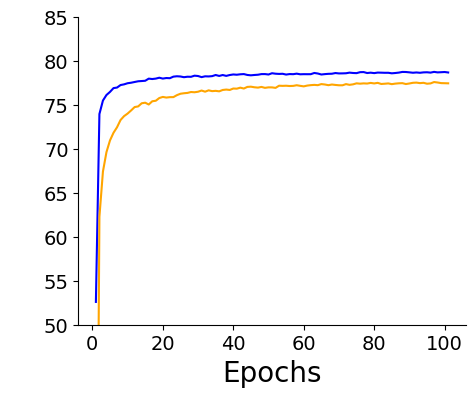}
        \label{fig:convergence-deit-s}
    }
    \subfigure[ViT-B]{
        \includegraphics[width=0.22\linewidth]{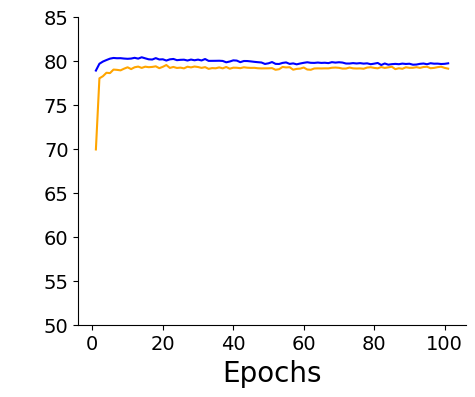}
        \label{fig:convergence-vit-b}
    }
     \subfigure[ViT-L]{
        \includegraphics[width=0.205\linewidth]{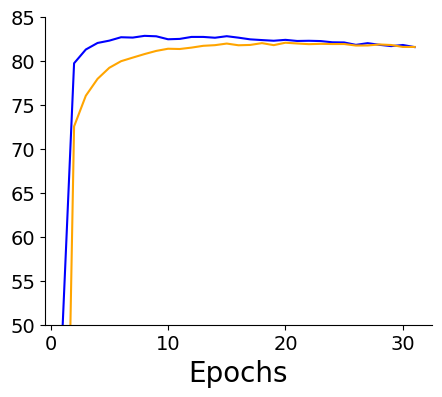}
        \label{fig:convergence-vit-l}
    }
   \caption{\textbf{\ouralgo achieves better performance and faster convergence on ImageNet-1k.} Finetuning with \ouralgo: Results comparing \ouralgo with standard routing and isoFLOP baselines with $50\%$ capacity on ImageNet.}
    \label{fig:convergence-in1k}
\end{figure}



\subsection{Faster convergence with \ouralgoheading on transfer learning}\label{sec:transfer-learning}
\begin{figure}
    \centering
    \subfigure[DeiT-T]{
        \includegraphics[width=0.22\linewidth]{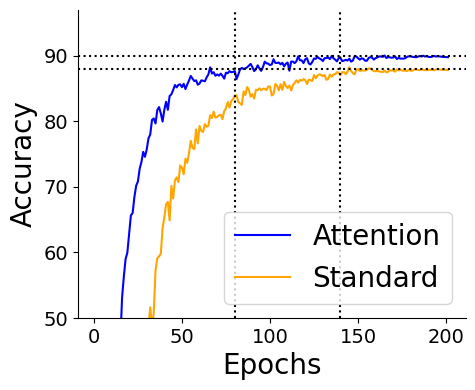}
        \label{fig:convergence-deit-t-flower}
    }
    \subfigure[DeiT-S]{
        \includegraphics[width=0.22\linewidth]{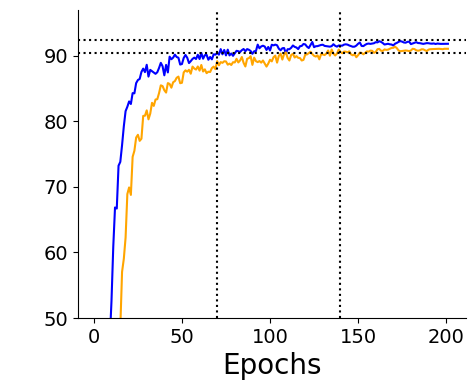}
        \label{fig:convergence-deit-s-flower}
    }
    \subfigure[ViT-B]{
        \includegraphics[width=0.22\linewidth]{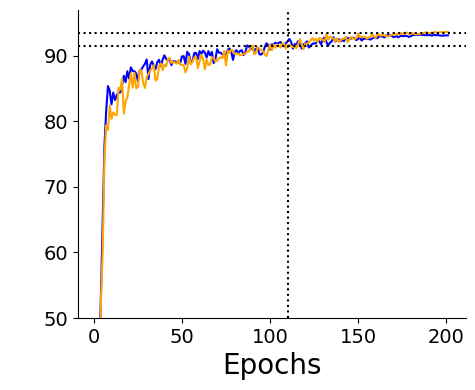}
        \label{fig:convergence-vit-b-flower}
    }
     \subfigure[ViT-L]{
        \includegraphics[width=0.20\linewidth]{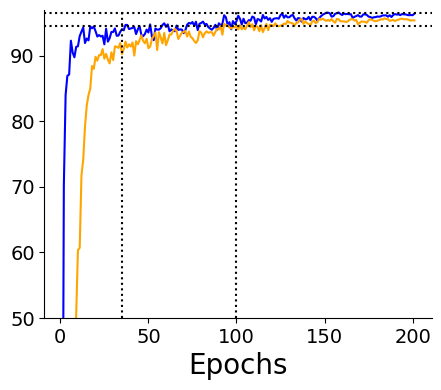}
        \label{fig:convergence-vit-l-flower}
    }
    
   \caption{\textbf{\ouralgo converges faster across different datasets} Transfer learning with \ouralgo: \ouralgo with $50\%$ capacity MoD trained on the Flower102 dataset. Dotted lines denote the epochs needed to reach within $2\%$ of peak accuracy.}
    \label{fig:convergence-transfer}
\end{figure}

We now investigate \ouralgo for transfer learning tasks from ImageNet-1k to three smaller image classification datasets: OxfordIIT-Pets, Stanford Cars and Flower102.
These tasks pose a challenge as the pretrained model must adapt to a MoD architecture with reduced capacity while training on limited data.
\cref{fig:convergence-transfer} reports the accuracy curves for \ouralgo in comparison with MoD on  Flower102 datasets.
Results for Stanford Cars and OxfordIIIT-Pets datasets are provided in \cref{fig:convergence-transfer-app} in the Appendix.

Across all datasets and model architectures, we find that \ouralgo converges faster in comparison to standard routing while outperforming standard routing in most cases.
We analyze convergence by measuring the number of epochs required for either model to reach within $2\%$ of its peak accuracy.
The black dotted lines in \cref{fig:convergence-transfer} and \cref{fig:convergence-transfer-app} help visualize this convergence for both \ouralgo and standard routing.
For ViT-Large on Flowers, (see \cref{fig:convergence-vit-l-flower}), \ouralgo reaches $94.5\%$ accuracy in the $35$-th epochs, while standard routing requires $100$-th epochs to reach the same value.
Similarly, in case of DeiT-S on Pets, (see \cref{fig:convergence-deit-s-pets} in the Appendix), \ouralgo reaches $90\%$ accuracy, in $25$ epochs while standard routing takes $70$ epochs to reach the same accuracy, enabling a $\sim2\times$ speed up.
These observations are consistent across architectures and datasets and highlight the effectiveness of \ouralgo to transfer MoD models from pretrained checkpoints.

\paragraph{isoFLOP comparison}
We also compare \ouralgo and standard routing to isoFLOP models on transfer learning tasks for both $50\%$ and $12.5\%$ capacity in \cref{tab:accuracy-transfer-12.5,tab:accuracy-transfer-50} in the appendix.
We find that MoD models are unable to match the isoFLOP model performance on transfer tasks.
We observe this as a limitation of the MoD framework in general for transfer learning on image tasks, irrespective of the routing mechanism employed. We propose a potential remedy to also outperform isoFLOP models in \cref{sec:mod-layer-position}. 

\subsection{Attention routing identifies important tokens}
\label{sec:token-importance}
To understand why \ouralgo improves over standard routing, we investigate the routing scores and their correlation with leave-one-out \citep{hastie2009elements} token importance.
Our goal is to estimate the relationship between the importance of a token and the routing score assigned to it by a standard or \ouralgo router.
Based on our empirical results, we conjecture that \ouralgo weights are better correlated with token importance in comparison with standard routing, thus enabling \ouralgo to always choose the most relevant tokens.

\begin{figure}
    \centering
    {\includegraphics[width=0.95\linewidth]{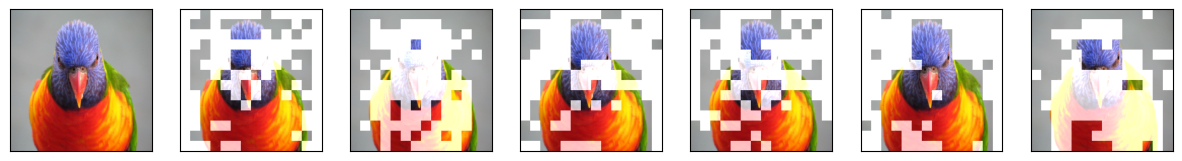}};
    {\includegraphics[width=0.95\linewidth]{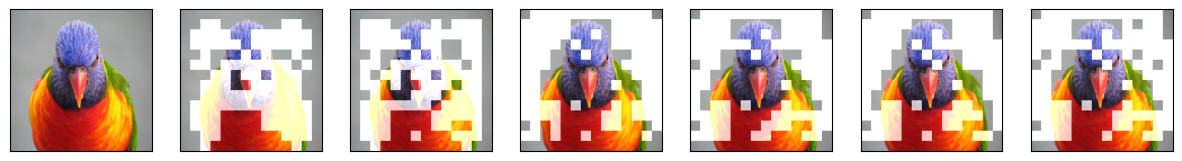}};
    \caption{\textbf{\ouralgo exhibits more meaningful routing compared to MoD.} Routing visualization: Example of DeiT-Small with $50\%$ capacity on ImageNet. Each example shows tokens chosen by standard MoD (top) and \ouralgo (bottom) for every MoD layer, white patches denote skipped. Each column represents a MoD layer as depth increases from left to right.}
    \label{fig:example-routing-deit-s-in1k}
\end{figure}

We first verify this claim by visualizing the routing in case of individual examples from ImageNet-1k as shown in \cref{fig:example-routing-deit-s-in1k}.
The figure highlights which patches of the image are chosen by the router in each MoD layer.
In case of \ouralgo (bottom),
the router selects tokens that are part of the bird outline and face starting from the third MoD layer.
In contrast, standard routing (top) selects more tokens that are part of the background, up to the last layer.

\begin{figure}
    \centering
    {\includegraphics[width=0.95\linewidth]{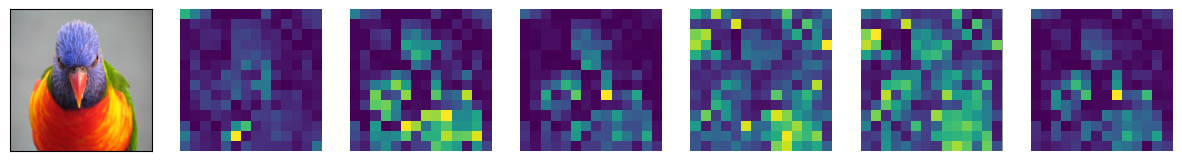}};
  {\includegraphics[width=0.95\linewidth]{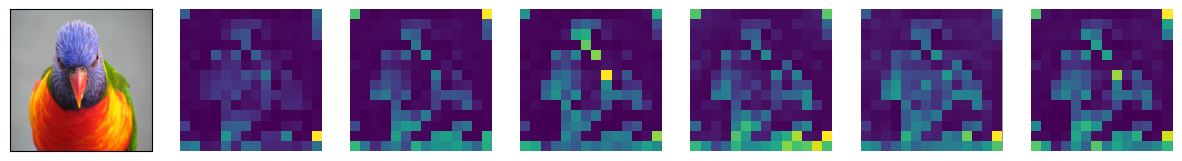}};

    \caption{\textbf{\ouralgo generates more meaningful attention maps compared to MoD.} Attention visualization: Example of DeiT-Small with $50\%$ capacity on ImageNet. The attention maps of the last MoD layer for standard routing (top row) and \ouralgo (bottom row) for each example. Each column denotes an attention head of the last layer.}
    \label{fig:example-attention-deit-s}
\end{figure}

Visualizing the attention maps of the last layer in \cref{fig:example-attention-deit-s} also confirms that \ouralgo is able to focus on the object in the image, which we use as routing scores.
The attention map for each head in the last layer for DeiT-Small identifies the silhouette of the bird for \ouralgo, but struggles for standard routing.
However, as shown by \citet{darcetvision}, attention maps do not always learn semantically meaningful scores.
This holds especially for larger models, where the attention scores tend to concentrate on a single patch (token) (see \cref{fig:example-attention-vit-b} in the Appendix).

    

\begin{figure}
    \centering
    \subfigure[DeiT-T]{
        \includegraphics[width=0.22\linewidth]{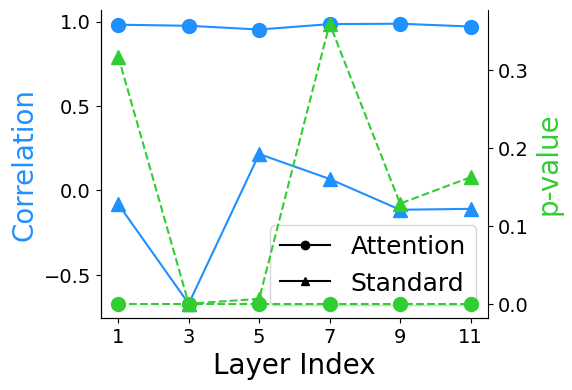}
        \label{fig:corr-deit-t}
    }
    \subfigure[DeiT-S]{
        \includegraphics[width=0.22\linewidth]{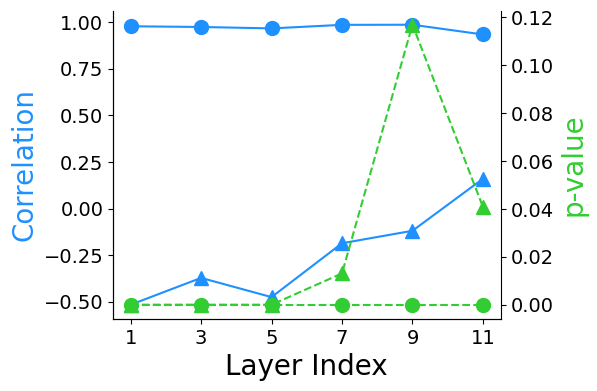}
        \label{fig:corr-deit-s}
    }
    \subfigure[ViT-B]{
        \includegraphics[width=0.22\linewidth]{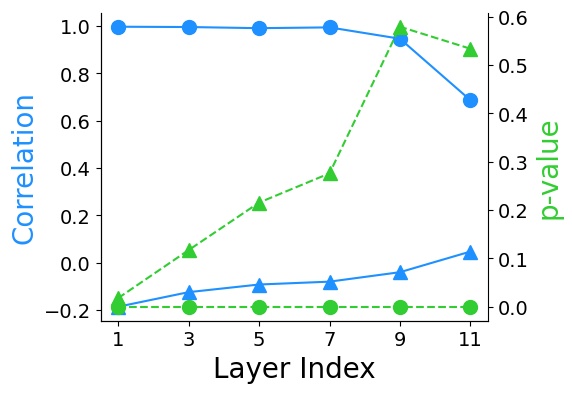}
        \label{fig:corr-vit-b}
    }
    \subfigure[ViT-L]{
        \includegraphics[width=0.22\linewidth]{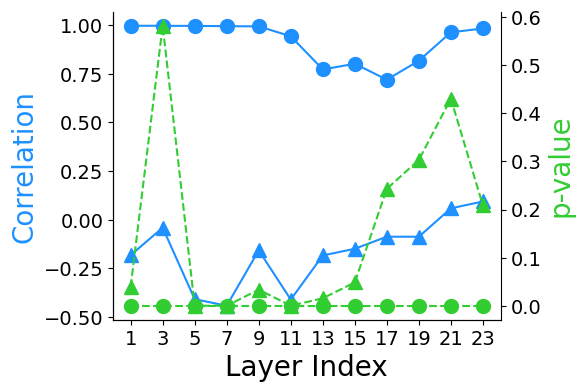}
        \label{fig:corr-vit-l}
    }
    \caption{\textbf{\ouralgo shows higher correlation between routing scores and leave-one-out token importance.} Correlation and p-values of the routing scores with layer-wise leave-one-out token importance on ImageNet.}
    \label{fig:routing-correlation}
\end{figure}

To quantify our qualitative observations, we compute the correlation of the routing scores with token importance estimates.
For the importance of a token, we compute the change in loss of the model if that token is omitted in the vanilla transformer i.e. leave-one-out token importance.
A large change in loss implies higher token importance and we would expect that token to have a higher routing score.
The correlation of the routing scores for both standard routing and \ouralgo with the token importance is shown in \cref{fig:routing-correlation} along with the corresponding p-values.

We observe that routing scores computed by \ouralgo consistently have a very high correlation with token importance suggesting that attention routing assigns higher scores to important tokens.
In contrast, standard routing sometimes even has a negative correlation with token importance, implying that it can assign higher scores to less important tokens.
Moreover, all the p-values observed for \ouralgo were lower than $10^{-8}$, whereas they were significant (in some layers even larger than $0.5$) in case of standard routing, implying higher uncertainty in case of standard routing.

\subsection{Impact of MoD in different layers}
\label{sec:mod-layer-position}
In our experiments so far, MoD layers are used in alternate layers following \citet{raposo2024mixture}. 
This model architecture gives us Pareto-optimal results on ImageNet-1k (see \cref{tab:accuracy-in1k}).
We conduct an ablation study to investigate whether introducing MoDs only in the later layers and keeping the initial layers dense is advantageous, particularly for visual tasks like classification, where learning low-level features may be critical.
In order to verify if MoDs benefit from additional feature learning at full capacity in the earlier layers, we introduce MoD layers alternately starting from the $4$-th layer, keeping the first four layers dense.

\begin{figure}
    \centering
    \subfigure[DeiT-S]{
        \includegraphics[width=0.40\linewidth]{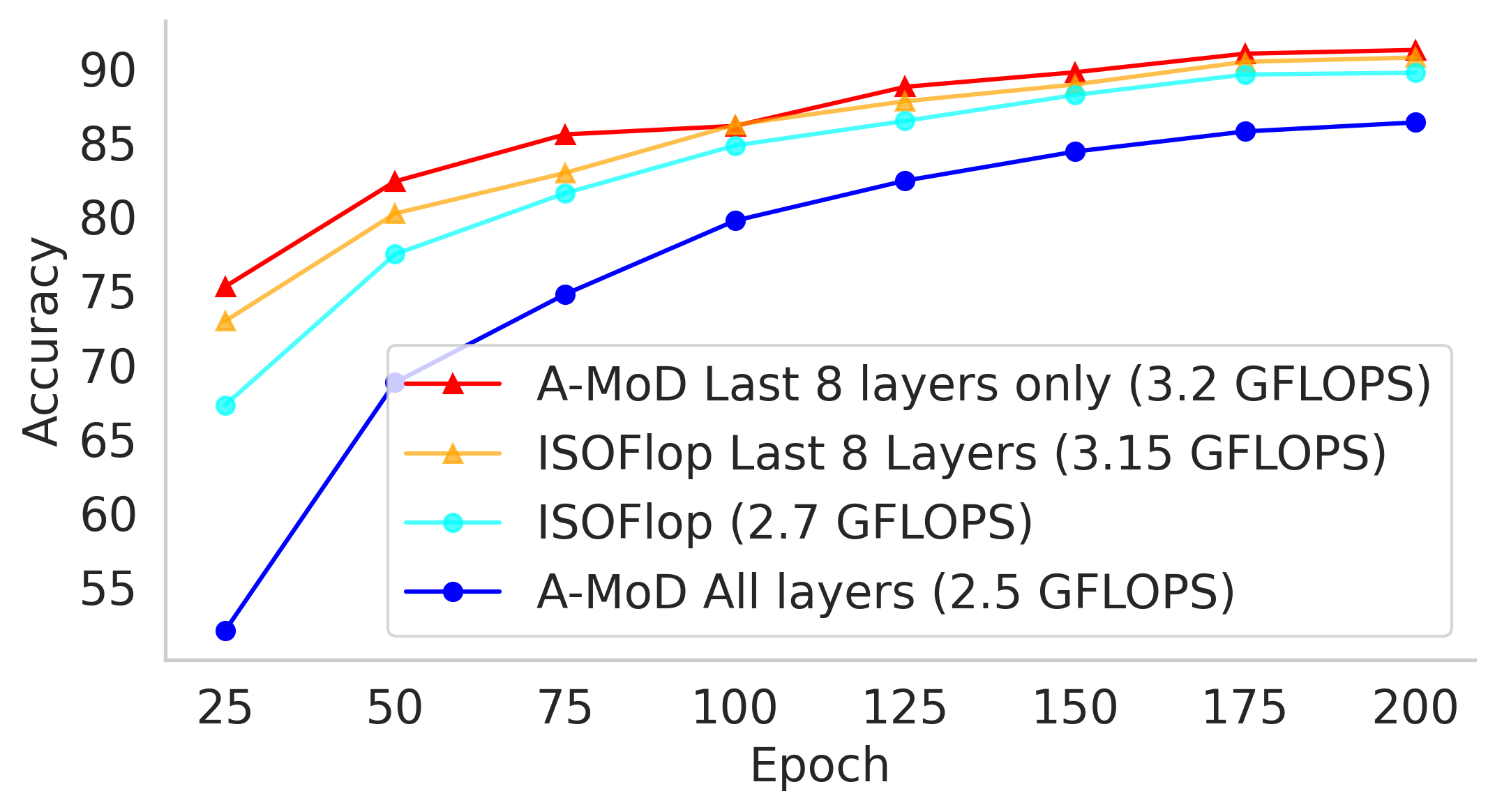}
        \label{fig:modlast8layers_deits}
    }
    \subfigure[ViT-B]{
        \includegraphics[width=0.40\linewidth]{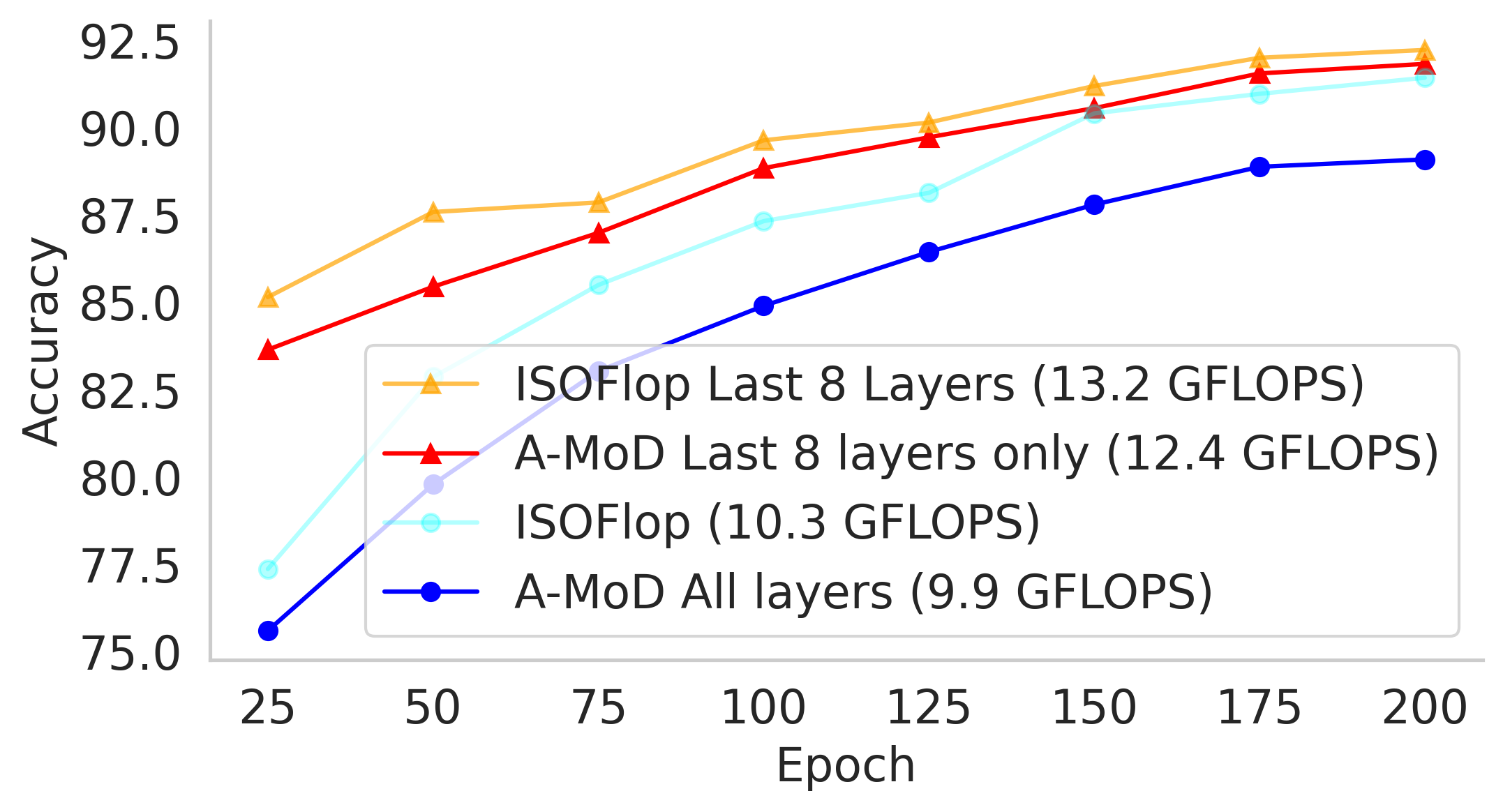}
        \label{fig:modlast8layers_vitb}
    }
    \caption{\textbf{\ouralgo improves the performance when only used in deeper layers.} 
    Introducing MoDs only in the last $8$ layers matches isoFLOP performance on the Stanford Cars dataset. 
    }
    \label{fig:modlast8layers_combined}
\end{figure}

Results in \cref{fig:modlast8layers_combined} show that keeping the first four layers dense improves on DeiT-Small and ViT-Base models trained on the Stanford Cars dataset.
The additional FLOPs allows for better learning in this regime as shown in \cref{fig:modlast8layers_combined}.
With this modification, \ouralgo is able to match the corresponding isoFLOP baseline, even for transfer learning tasks.
This highlights a potential method to address the limitations of \ouralgo mentioned in \cref{sec:transfer-learning} at the cost of additional FLOPs. 

\section{Conclusion}

We propose \ouralgo, a variation of Mixture-of-Depths (MoD) with attention routing instead of a standard router. To compute token importance for an MoD layer, \ouralgo utilizes the attention maps from its previous layer, thereby achieving attention routing without additional parameters. In case of training from a pretrained checkpoint, leveraging trained attention information also leads to increased training stability and faster convergence compared to vanilla MoD. Furthermore, we empirically demonstrate that \ouralgo outperforms standard MoD across different model configurations and datasets while making better routing decisions.



\newpage


\begin{thebibliography}{47}
\providecommand{\natexlab}[1]{#1}
\providecommand{\url}[1]{\texttt{#1}}
\expandafter\ifx\csname urlstyle\endcsname\relax
  \providecommand{\doi}[1]{doi: #1}\else
  \providecommand{\doi}{doi: \begingroup \urlstyle{rm}\Url}\fi

\bibitem[Bahdanau(2014)]{bahdanau2014neural}
Dzmitry Bahdanau.
\newblock Neural machine translation by jointly learning to align and translate.
\newblock \emph{arXiv preprint arXiv:1409.0473}, 2014.

\bibitem[Bengio et~al.(2015)Bengio, Bacon, Pineau, and Precup]{bengio2015conditional}
Emmanuel Bengio, Pierre-Luc Bacon, Joelle Pineau, and Doina Precup.
\newblock Conditional computation in neural networks for faster models.
\newblock \emph{arXiv preprint arXiv:1511.06297}, 2015.

\bibitem[Bengio et~al.(2013)Bengio, L{\'e}onard, and Courville]{bengio2013estimating}
Yoshua Bengio, Nicholas L{\'e}onard, and Aaron Courville.
\newblock Estimating or propagating gradients through stochastic neurons for conditional computation.
\newblock \emph{arXiv preprint arXiv:1308.3432}, 2013.

\bibitem[Bolya et~al.()Bolya, Fu, Dai, Zhang, Feichtenhofer, and Hoffman]{tome}
Daniel Bolya, Cheng-Yang Fu, Xiaoliang Dai, Peizhao Zhang, Christoph Feichtenhofer, and Judy Hoffman.
\newblock Token merging: Your vit but faster.
\newblock In \emph{The Eleventh International Conference on Learning Representations}.

\bibitem[Cai et~al.(2024)Cai, Jiang, Wang, Tang, Kim, and Huang]{cai2024survey}
Weilin Cai, Juyong Jiang, Fan Wang, Jing Tang, Sunghun Kim, and Jiayi Huang.
\newblock A survey on mixture of experts.
\newblock \emph{arXiv preprint arXiv:2407.06204}, 2024.

\bibitem[Carion et~al.(2020{\natexlab{a}})Carion, Massa, Synnaeve, Usunier, Kirillov, and Zagoruyko]{carion2020end}
Nicolas Carion, Francisco Massa, Gabriel Synnaeve, Nicolas Usunier, Alexander Kirillov, and Sergey Zagoruyko.
\newblock End-to-end object detection with transformers.
\newblock In \emph{European conference on computer vision}, pp.\  213--229. Springer, 2020{\natexlab{a}}.

\bibitem[Carion et~al.(2020{\natexlab{b}})Carion, Massa, Synnaeve, Usunier, Kirillov, and Zagoruyko]{detr}
Nicolas Carion, Francisco Massa, Gabriel Synnaeve, Nicolas Usunier, Alexander Kirillov, and Sergey Zagoruyko.
\newblock End-to-end object detection with transformers.
\newblock In \emph{European conference on computer vision}, pp.\  213--229. Springer, 2020{\natexlab{b}}.

\bibitem[Clark et~al.(2022)Clark, de~Las~Casas, Guy, Mensch, Paganini, Hoffmann, Damoc, Hechtman, Cai, Borgeaud, et~al.]{clark2022unified}
Aidan Clark, Diego de~Las~Casas, Aurelia Guy, Arthur Mensch, Michela Paganini, Jordan Hoffmann, Bogdan Damoc, Blake Hechtman, Trevor Cai, Sebastian Borgeaud, et~al.
\newblock Unified scaling laws for routed language models.
\newblock In \emph{International conference on machine learning}, pp.\  4057--4086. PMLR, 2022.

\bibitem[Dao et~al.(2022)Dao, Fu, Ermon, Rudra, and R{\'e}]{dao2022flashattention}
Tri Dao, Dan Fu, Stefano Ermon, Atri Rudra, and Christopher R{\'e}.
\newblock Flashattention: Fast and memory-efficient exact attention with io-awareness.
\newblock \emph{Advances in Neural Information Processing Systems}, 35:\penalty0 16344--16359, 2022.

\bibitem[Darcet et~al.(2024)Darcet, Oquab, Mairal, and Bojanowski]{darcetvision}
Timoth{\'e}e Darcet, Maxime Oquab, Julien Mairal, and Piotr Bojanowski.
\newblock Vision transformers need registers.
\newblock In \emph{International Conference on Learning Representations}, 2024.

\bibitem[Dikkala et~al.(2023)Dikkala, Ghosh, Meka, Panigrahy, Vyas, and Wang]{dikkala2023benefits}
Nishanth Dikkala, Nikhil Ghosh, Raghu Meka, Rina Panigrahy, Nikhil Vyas, and Xin Wang.
\newblock On the benefits of learning to route in mixture-of-experts models.
\newblock In \emph{Proceedings of the 2023 Conference on Empirical Methods in Natural Language Processing}, pp.\  9376--9396, 2023.

\bibitem[Dosovitskiy et~al.(2021)Dosovitskiy, Beyer, Kolesnikov, Weissenborn, Zhai, Unterthiner, Dehghani, Minderer, Heigold, Gelly, Uszkoreit, and Houlsby]{dosovitskiy2021an}
Alexey Dosovitskiy, Lucas Beyer, Alexander Kolesnikov, Dirk Weissenborn, Xiaohua Zhai, Thomas Unterthiner, Mostafa Dehghani, Matthias Minderer, Georg Heigold, Sylvain Gelly, Jakob Uszkoreit, and Neil Houlsby.
\newblock An image is worth 16x16 words: Transformers for image recognition at scale.
\newblock In \emph{International Conference on Learning Representations}, 2021.
\newblock URL \url{https://openreview.net/forum?id=YicbFdNTTy}.

\bibitem[Fedus et~al.(2022{\natexlab{a}})Fedus, Dean, and Zoph]{fedus2022review}
William Fedus, Jeff Dean, and Barret Zoph.
\newblock A review of sparse expert models in deep learning.
\newblock \emph{arXiv preprint arXiv:2209.01667}, 2022{\natexlab{a}}.

\bibitem[Fedus et~al.(2022{\natexlab{b}})Fedus, Zoph, and Shazeer]{fedus2022switch}
William Fedus, Barret Zoph, and Noam Shazeer.
\newblock Switch transformers: Scaling to trillion parameter models with simple and efficient sparsity.
\newblock \emph{Journal of Machine Learning Research}, 23\penalty0 (120):\penalty0 1--39, 2022{\natexlab{b}}.

\bibitem[Hastie et~al.(2009)Hastie, Tibshirani, Friedman, and Friedman]{hastie2009elements}
Trevor Hastie, Robert Tibshirani, Jerome~H Friedman, and Jerome~H Friedman.
\newblock \emph{The elements of statistical learning: data mining, inference, and prediction}, volume~2.
\newblock Springer, 2009.

\bibitem[He et~al.(2016)He, Zhang, Ren, and Sun]{he2016deep}
Kaiming He, Xiangyu Zhang, Shaoqing Ren, and Jian Sun.
\newblock Deep residual learning for image recognition.
\newblock In \emph{Proceedings of the IEEE conference on computer vision and pattern recognition}, pp.\  770--778, 2016.

\bibitem[Hoffmann et~al.(2022)Hoffmann, Borgeaud, Mensch, Buchatskaya, Cai, Rutherford, Casas, Hendricks, Welbl, Clark, et~al.]{hoffmann2022training}
Jordan Hoffmann, Sebastian Borgeaud, Arthur Mensch, Elena Buchatskaya, Trevor Cai, Eliza Rutherford, Diego de~Las Casas, Lisa~Anne Hendricks, Johannes Welbl, Aidan Clark, et~al.
\newblock Training compute-optimal large language models.
\newblock \emph{arXiv preprint arXiv:2203.15556}, 2022.

\bibitem[Jacobs et~al.(1991)Jacobs, Jordan, Nowlan, and Hinton]{jacobs1991adaptive}
Robert~A Jacobs, Michael~I Jordan, Steven~J Nowlan, and Geoffrey~E Hinton.
\newblock Adaptive mixtures of local experts.
\newblock \emph{Neural computation}, 3\penalty0 (1):\penalty0 79--87, 1991.

\bibitem[Jain et~al.(2024)Jain, Hegde, Kusupati, Nagrani, Buch, Jain, Arnab, and Paul]{jain2024mixture}
Gagan Jain, Nidhi Hegde, Aditya Kusupati, Arsha Nagrani, Shyamal Buch, Prateek Jain, Anurag Arnab, and Sujoy Paul.
\newblock Mixture of nested experts: Adaptive processing of visual tokens.
\newblock \emph{arXiv preprint arXiv:2407.19985}, 2024.

\bibitem[Jetley et~al.(2018)Jetley, Lord, Lee, and Torr]{jetley2018learn}
Saumya Jetley, Nicholas~A Lord, Namhoon Lee, and Philip~HS Torr.
\newblock Learn to pay attention.
\newblock In \emph{International Conference on Learning Representations}, 2018.

\bibitem[Jiang et~al.(2024)Jiang, Sablayrolles, Roux, Mensch, Savary, Bamford, Chaplot, Casas, Hanna, Bressand, et~al.]{jiang2024mixtral}
Albert~Q Jiang, Alexandre Sablayrolles, Antoine Roux, Arthur Mensch, Blanche Savary, Chris Bamford, Devendra~Singh Chaplot, Diego de~las Casas, Emma~Bou Hanna, Florian Bressand, et~al.
\newblock Mixtral of experts.
\newblock \emph{arXiv preprint arXiv:2401.04088}, 2024.

\bibitem[Jordan \& Jacobs(1993)Jordan and Jacobs]{og_moe}
M.I. Jordan and R.A. Jacobs.
\newblock Hierarchical mixtures of experts and the {EM} algorithm.
\newblock In \emph{Proceedings of 1993 International Conference on Neural Networks (IJCNN-93-Nagoya, Japan)}, volume~2, pp.\  1339--1344 vol.2, 1993.
\newblock \doi{10.1109/IJCNN.1993.716791}.

\bibitem[Kaplan et~al.(2020)Kaplan, McCandlish, Henighan, Brown, Chess, Child, Gray, Radford, Wu, and Amodei]{kaplan2020scaling}
Jared Kaplan, Sam McCandlish, Tom Henighan, Tom~B Brown, Benjamin Chess, Rewon Child, Scott Gray, Alec Radford, Jeffrey Wu, and Dario Amodei.
\newblock Scaling laws for neural language models.
\newblock \emph{arXiv preprint arXiv:2001.08361}, 2020.

\bibitem[Krause et~al.(2013)Krause, Deng, Stark, and Fei-Fei]{cars}
Jonathan Krause, Jia Deng, Michael Stark, and Li~Fei-Fei.
\newblock Collecting a large-scale dataset of fine-grained cars, 2013.

\bibitem[Lewis et~al.(2021)Lewis, Bhosale, Dettmers, Goyal, and Zettlemoyer]{lewis2021base}
Mike Lewis, Shruti Bhosale, Tim Dettmers, Naman Goyal, and Luke Zettlemoyer.
\newblock Base layers: Simplifying training of large, sparse models.
\newblock In \emph{International Conference on Machine Learning}, pp.\  6265--6274. PMLR, 2021.

\bibitem[Liang et~al.(2022)Liang, Chongjian, Tong, Song, Wang, and Xie]{liangevit}
Youwei Liang, GE~Chongjian, Zhan Tong, Yibing Song, Jue Wang, and Pengtao Xie.
\newblock {EViT}: Expediting vision transformers via token reorganizations.
\newblock In \emph{International Conference on Learning Representations}, 2022.

\bibitem[Liu et~al.(2024)Liu, Blondel, Ruiz, and Puigcerver]{liurouters}
Tianlin Liu, Mathieu Blondel, Carlos~Riquelme Ruiz, and Joan Puigcerver.
\newblock Routers in vision mixture of experts: An empirical study.
\newblock \emph{Transactions on Machine Learning Research}, 2024.

\bibitem[Loshchilov \& Hutter(2017)Loshchilov and Hutter]{adamw}
I~Loshchilov and F~Hutter.
\newblock Decoupled weight decay regularization.
\newblock \emph{arXiv preprint arXiv:1711.05101}, 2017.

\bibitem[Ludziejewski et~al.(2024)Ludziejewski, Krajewski, Adamczewski, Pi{\'o}ro, Krutul, Antoniak, Ciebiera, Kr{\'o}l, Odrzyg{\'o}{\'z}d{\'z}, Sankowski, et~al.]{ludziejewskiscaling}
Jan Ludziejewski, Jakub Krajewski, Kamil Adamczewski, Maciej Pi{\'o}ro, Micha{\l} Krutul, Szymon Antoniak, Kamil Ciebiera, Krystian Kr{\'o}l, Tomasz Odrzyg{\'o}{\'z}d{\'z}, Piotr Sankowski, et~al.
\newblock Scaling laws for fine-grained mixture of experts.
\newblock In \emph{ICLR 2024 Workshop on Mathematical and Empirical Understanding of Foundation Models}, 2024.

\bibitem[Nilsback \& Zisserman(2008)Nilsback and Zisserman]{flowers}
Maria-Elena Nilsback and Andrew Zisserman.
\newblock Automated flower classification over a large number of classes.
\newblock In \emph{Indian Conference on Computer Vision, Graphics and Image Processing}, Dec 2008.

\bibitem[Parkhi et~al.(2012)Parkhi, Vedaldi, Zisserman, and Jawahar]{pets}
Omkar~M. Parkhi, Andrea Vedaldi, Andrew Zisserman, and C.~V. Jawahar.
\newblock Cats and dogs.
\newblock In \emph{IEEE Conference on Computer Vision and Pattern Recognition}, 2012.

\bibitem[Puigcerver et~al.(2024)Puigcerver, Ruiz, Mustafa, and Houlsby]{puigcerversparse}
Joan Puigcerver, Carlos~Riquelme Ruiz, Basil Mustafa, and Neil Houlsby.
\newblock From sparse to soft mixtures of experts.
\newblock In \emph{International Conference on Learning Representations}, 2024.

\bibitem[Ramachandran \& Le(2019)Ramachandran and Le]{ramachandran2018diversity}
Prajit Ramachandran and Quoc~V. Le.
\newblock Diversity and depth in per-example routing models.
\newblock In \emph{International Conference on Learning Representations}, 2019.
\newblock URL \url{https://openreview.net/forum?id=BkxWJnC9tX}.

\bibitem[Rao et~al.(2021)Rao, Zhao, Liu, Lu, Zhou, and Hsieh]{rao2021dynamicvit}
Yongming Rao, Wenliang Zhao, Benlin Liu, Jiwen Lu, Jie Zhou, and Cho-Jui Hsieh.
\newblock Dynamicvit: Efficient vision transformers with dynamic token sparsification.
\newblock \emph{Advances in neural information processing systems}, 34:\penalty0 13937--13949, 2021.

\bibitem[Raposo et~al.(2024)Raposo, Ritter, Richards, Lillicrap, Humphreys, and Santoro]{raposo2024mixture}
David Raposo, Sam Ritter, Blake Richards, Timothy Lillicrap, Peter~Conway Humphreys, and Adam Santoro.
\newblock Mixture-of-depths: Dynamically allocating compute in transformer-based language models.
\newblock \emph{arXiv preprint arXiv:2404.02258}, 2024.

\bibitem[Riquelme et~al.(2021)Riquelme, Puigcerver, Mustafa, Neumann, Jenatton, Susano~Pinto, Keysers, and Houlsby]{riquelme2021scaling}
Carlos Riquelme, Joan Puigcerver, Basil Mustafa, Maxim Neumann, Rodolphe Jenatton, Andr{\'e} Susano~Pinto, Daniel Keysers, and Neil Houlsby.
\newblock Scaling vision with sparse mixture of experts.
\newblock \emph{Advances in Neural Information Processing Systems}, 34:\penalty0 8583--8595, 2021.

\bibitem[Roller et~al.(2021)Roller, Sukhbaatar, Weston, et~al.]{roller2021hash}
Stephen Roller, Sainbayar Sukhbaatar, Jason Weston, et~al.
\newblock Hash layers for large sparse models.
\newblock \emph{Advances in Neural Information Processing Systems}, 34:\penalty0 17555--17566, 2021.

\bibitem[Russakovsky et~al.(2015)Russakovsky, Deng, Su, Krause, Satheesh, Ma, Huang, Karpathy, Khosla, Bernstein, Berg, and Fei-Fei]{imagenet}
Olga Russakovsky, Jia Deng, Hao Su, Jonathan Krause, Sanjeev Satheesh, Sean Ma, Zhiheng Huang, Andrej Karpathy, Aditya Khosla, Michael Bernstein, Alexander~C. Berg, and Li~Fei-Fei.
\newblock {ImageNet Large Scale Visual Recognition Challenge}.
\newblock \emph{International Journal of Computer Vision (IJCV)}, 115\penalty0 (3):\penalty0 211--252, 2015.
\newblock \doi{10.1007/s11263-015-0816-y}.

\bibitem[Shazeer et~al.(2016)Shazeer, Mirhoseini, Maziarz, Davis, Le, Hinton, and Dean]{shazeer2016outrageously}
Noam Shazeer, Azalia Mirhoseini, Krzysztof Maziarz, Andy Davis, Quoc Le, Geoffrey Hinton, and Jeff Dean.
\newblock Outrageously large neural networks: The sparsely-gated mixture-of-experts layer.
\newblock In \emph{International Conference on Learning Representations}, 2016.

\bibitem[Thompson et~al.(2020)Thompson, Greenewald, Lee, and Manso]{thompson2020computational}
Neil~C Thompson, Kristjan Greenewald, Keeheon Lee, and Gabriel~F Manso.
\newblock The computational limits of deep learning.
\newblock \emph{arXiv preprint arXiv:2007.05558}, 2020.

\bibitem[Touvron et~al.(2021)Touvron, Cord, Douze, Massa, Sablayrolles, and Jegou]{deit}
Hugo Touvron, Matthieu Cord, Matthijs Douze, Francisco Massa, Alexandre Sablayrolles, and Herve Jegou.
\newblock Training data-efficient image transformers \& distillation through attention.
\newblock In Marina Meila and Tong Zhang (eds.), \emph{Proceedings of the 38th International Conference on Machine Learning}, volume 139 of \emph{Proceedings of Machine Learning Research}, pp.\  10347--10357. PMLR, 18--24 Jul 2021.
\newblock URL \url{https://proceedings.mlr.press/v139/touvron21a.html}.

\bibitem[Vaswani et~al.(2017)Vaswani, Shazeer, Parmar, Uszkoreit, Jones, Gomez, Kaiser, and Polosukhin]{vaswani2017attention}
Ashish Vaswani, Noam Shazeer, Niki Parmar, Jakob Uszkoreit, Llion Jones, Aidan~N Gomez, Lukasz Kaiser, and Illia Polosukhin.
\newblock Attention is all you need.
\newblock \emph{Advances in Neural Information Processing Systems}, 2017.

\bibitem[Wang et~al.(2024)Wang, Chen, Liu, Chen, Lin, Han, and Ding]{wang2024yolov10}
Ao~Wang, Hui Chen, Lihao Liu, Kai Chen, Zijia Lin, Jungong Han, and Guiguang Ding.
\newblock Yolov10: Real-time end-to-end object detection.
\newblock \emph{arXiv preprint arXiv:2405.14458}, 2024.

\bibitem[Wei et~al.(2022)Wei, Tay, Bommasani, Raffel, Zoph, Borgeaud, Yogatama, Bosma, Zhou, Metzler, et~al.]{wei2022emergent}
Jason Wei, Yi~Tay, Rishi Bommasani, Colin Raffel, Barret Zoph, Sebastian Borgeaud, Dani Yogatama, Maarten Bosma, Denny Zhou, Donald Metzler, et~al.
\newblock Emergent abilities of large language models.
\newblock \emph{arXiv preprint arXiv:2206.07682}, 2022.

\bibitem[Yin et~al.(2022)Yin, Vahdat, Alvarez, Mallya, Kautz, and Molchanov]{yin2022vit}
Hongxu Yin, Arash Vahdat, Jose~M Alvarez, Arun Mallya, Jan Kautz, and Pavlo Molchanov.
\newblock A-vit: Adaptive tokens for efficient vision transformer.
\newblock In \emph{Proceedings of the IEEE/CVF conference on computer vision and pattern recognition}, pp.\  10809--10818, 2022.

\bibitem[Zhou et~al.(2022)Zhou, Lei, Liu, Du, Huang, Zhao, Dai, Le, Laudon, et~al.]{zhou2022mixture}
Yanqi Zhou, Tao Lei, Hanxiao Liu, Nan Du, Yanping Huang, Vincent Zhao, Andrew~M Dai, Quoc~V Le, James Laudon, et~al.
\newblock Mixture-of-experts with expert choice routing.
\newblock \emph{Advances in Neural Information Processing Systems}, 35:\penalty0 7103--7114, 2022.

\bibitem[Zoph et~al.(2022)Zoph, Bello, Kumar, Du, Huang, Dean, Shazeer, and Fedus]{zoph2202st}
Barret Zoph, Irwan Bello, Sameer Kumar, Nan Du, Yanping Huang, Jeff Dean, Noam Shazeer, and William Fedus.
\newblock {ST-MoE}: Designing stable and transferable sparse expert models.
\newblock \emph{arXiv preprint arXiv:2202.08906}, 2022.

\end{thebibliography}

\appendix
\newpage
\section{Appendix}

\subsection{Model Specs}
We choose four different transformer-based architectures: \\
\begin{table}[ht!]
    \centering
    \caption{Specifications of transformer-based models used in experiments}
    \label{tab:model_specs}
    \begin{tabular}{l c c}
        \textbf{Model} & \textbf{Parameters (M)} & \textbf{FLOPS (G)} \\
        DeiT-Tiny    & \hphantom{00}5.72    & \hphantom{00}1.26  \\
        DeiT-Small   & \hphantom{0}22.05   & \hphantom{00}4.61  \\
        ViT-Base     & \hphantom{0}86.57   & \hphantom{0}17.58   \\
        ViT-Large    & 304.72   & 191.21   \\
    \end{tabular}
\end{table}

\subsection{Datasets}
We evaluate our models on standard benchmark datasets: ImageNet-1K, Stanford Cars, Oxford Pets and Flowers.
We use the standard train and test splits for each dataset.
\begin{itemize}
    \item \textbf{ImageNet-1k}: A large-scale dataset comprising over 1.2 million images across 1,000 diverse categories, widely used for benchmarking image classification models \citep{imagenet}.
    \item \textbf{Stanford Cars}: Consists of 16,185 high-resolution images of 196 car categories, focusing on fine-grained classification tasks \citep{cars}.
    \item \textbf{Pets}: Contains 37 categories of Pets with approximately 200 images per class \citep{pets}.
    \item \textbf{Flowers}: Consists of 8,189 images of 102 flower categories, focusing on fine-grained classification tasks \citep{flowers}.
\end{itemize}



\subsection{Additional results for finetuning on ImageNet-1k}

\begin{figure}
    \centering
 
    \subfigure[DeiT-T]{ 
        \includegraphics[width=0.22\linewidth]{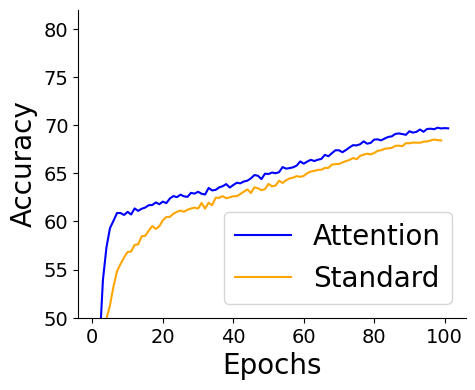}
        \label{fig:convergence-deit-t-125}
    }
    \subfigure[DeiT-S]{ 
        \includegraphics[width=0.22\linewidth]{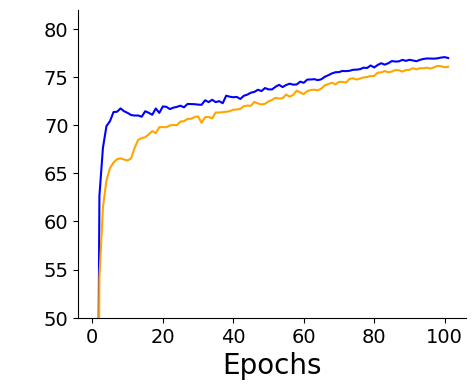}
        \label{fig:convergence-deit-s-125}
    }
    \subfigure[ViT-B]{ 
        \includegraphics[width=0.22\linewidth]{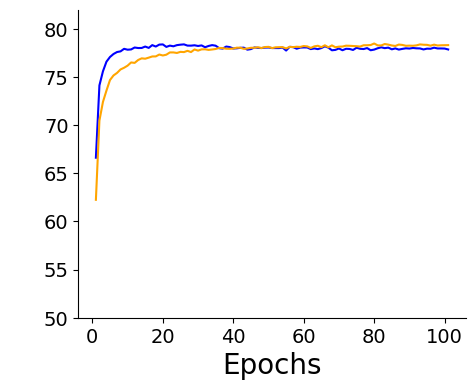}
        \label{fig:convergence-vit-b-125}
    }
     \subfigure[ViT-L]{
        \includegraphics[width=0.22\linewidth]{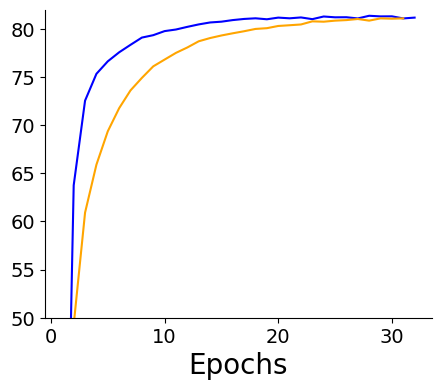}
        \label{fig:convergence-vit-l-125}
    }
   \caption{\textbf{\ouralgo achieves better performance and faster convergence on ImageNet-1k.} Finetuning with \ouralgo: Results comparing \ouralgo with standard routing and isoFLOP baselines for  $12.5\%$ capacity on ImageNet-1k.}
    \label{fig:convergence-in1k-app}
\end{figure}

\cref{fig:convergence-in1k-app} presents the convergence results for finetuning on ImageNet-1k with \ouralgo at $12.5\%$ capacity.
\cref{tab:accuracy-adapt} denotes the accuracy of \ouralgo and standard routing without any training, after adapting the MoD weights from a vanilla pretrained checkpoint.

\begin{table}[ht]
    \centering
    \caption{\textbf{\ouralgo improves adaptation.} Accuracy of MoD on ImageNet-1k, adapted from a pretrained checkpoint, without any training.}
    \label{tab:accuracy-adapt}
    \begin{tabular}{l c c c}
        \toprule
        \textbf{Model} & \textbf{Configuration} & \textbf{C = $50\%$} &  \textbf{C = $12.5\%$}  \\
        \midrule
       {DeiT-Tiny} & MoD  & 4.45 & 0.42\\
                                   & \ouralgo & \textbf{52.6} & \textbf{0.97} \\
        \midrule
     {DeiT-Small} & MoD  & 0.23 & 0.16\\
                                   & \ouralgo & \textbf{13.49} &\textbf{0.35} \\
        \midrule
       {ViT-Base}  & MoD  & 69.91 & 62.25\\
                                   & \ouralgo & \textbf{78.88} &\textbf{66.62} \\
       \midrule
  {ViT-Large} & MoD  & 0.43 & 0.2\\
                                   & \ouralgo &\textbf{49.06} & \textbf{6.03} \\
        \bottomrule
    \end{tabular}
\end{table}

\subsection{Comparison of \ouralgoheading with standard routing and isoFLOP baselines for transfer learning}

\begin{figure}
    \centering
      \subfigure[DeiT-T]{
        \includegraphics[width=0.22\linewidth]{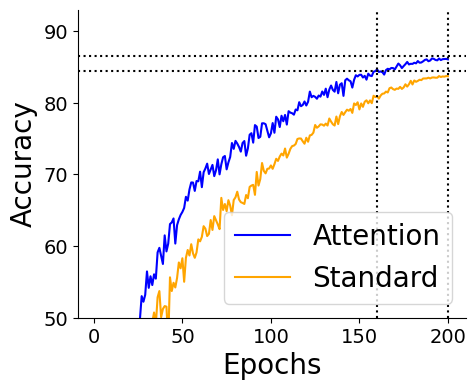}
        \label{fig:convergence-deit-t-cars}
    }
    \subfigure[DeiT-S]{
        \includegraphics[width=0.22\linewidth]{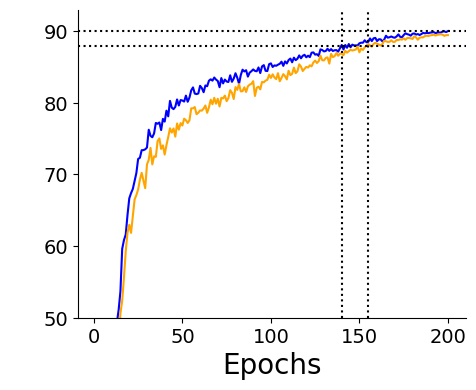}
        \label{fig:convergence-deit-s-cars}
    }
    \subfigure[ViT-B]{
        \includegraphics[width=0.22\linewidth]{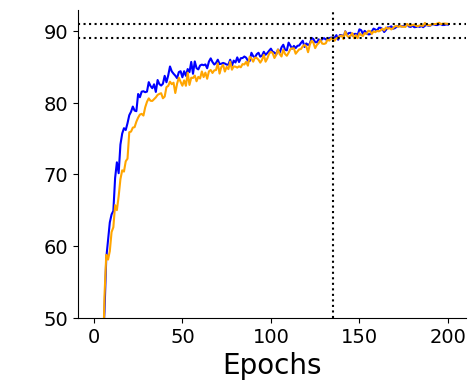}
        \label{fig:convergence-vit-b-cars}
    }
     \subfigure[ViT-L]{
        \includegraphics[width=0.22\linewidth]{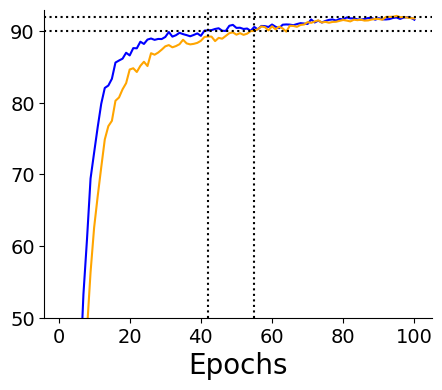}
        \label{fig:convergence-vit-l-cars}
    }
    \subfigure[DeiT-T]{
        \includegraphics[width=0.22\linewidth]{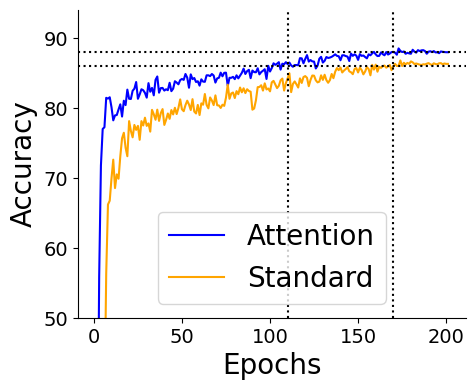}
        \label{fig:convergence-deit-t-pets}
    }
    \subfigure[DeiT-S]{
        \includegraphics[width=0.22\linewidth]{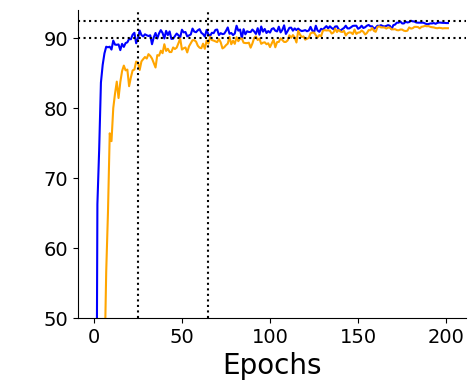}
        \label{fig:convergence-deit-s-pets}
    }
    \subfigure[ViT-B]{
        \includegraphics[width=0.22\linewidth]{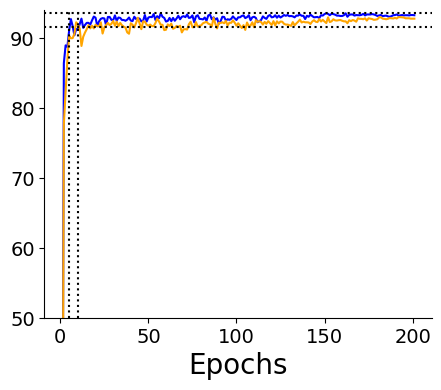}
        \label{fig:convergence-vit-b-pets}
    }
     \subfigure[ViT-L]{
        \includegraphics[width=0.22\linewidth]{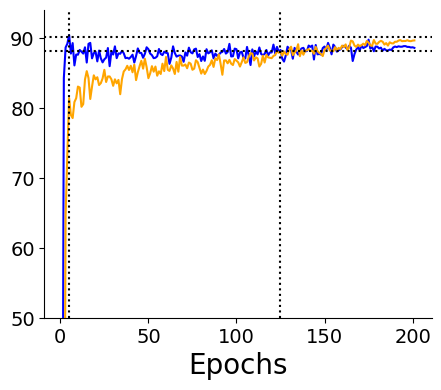}
        \label{fig:convergence-vit-l-pets}
    }
    
   \caption{\textbf{\ouralgo converges faster across different datasets} Transfer learning with \ouralgo: \ouralgo with $50\%$ capacity MoD trained on the Stanford Cars (top row) and OxfordIIT-Pets (bottom row) datasets. Dotted lines denote the epochs needed to reach within $2\%$ of peak accuracy.}
    \label{fig:convergence-transfer-app}
\end{figure}

Table \ref{tab:accuracy-transfer-12.5} and Table \ref{tab:accuracy-transfer-50} present the classification accuracy of each model configuration across the three datasets for $12.5\%$ and $50\%$ capacity respectively.
\cref{fig:convergence-transfer-app} presents the convergence results for \ouralgo on transfer learnign tasks on the Stanford Cars and OxfordIIT-Pets datasets.

\begin{table}[ht]
    \centering
    \caption{Top-1 accuracy for MoD models with $12.5\%$ capacity, compared with isoFLOP baselines.}
    \label{tab:accuracy-transfer-12.5}
    \begin{tabular}{l l c c c c }
        \toprule
        \textbf{Model} & \textbf{FLOPS (G)}  & \textbf{Configuration} & \textbf{Cars} & \textbf{Flowers} & \textbf{Pets} \\
        \midrule
        \multirow{4}{*}{DeiT-Tiny} & 0.746  & isoFLOP                       &  \textbf{85.09}           & \textbf{90.73}    &  \textbf{85.3}   \\
                                   & 0.709 & MoD   & 75.77             & 81.39   &  84.05   \\
                                   & 0.709 & \ouralgo  & 78.32  & 82.82  & 84.68\\
        \midrule
        \multirow{4}{*}{DeiT-Small} & 2.333 & isoFLOP                       & \textbf{89.88}         & \textbf{90.97}   & 86.42   \\
                                   & 2.592 & MoD   & 85.36             & 88.46    &   86.04 \\
                                   & 2.591  & \ouralgo  & 86.39  & 89.44 & \textbf{89.58} \\
        \midrule
        \multirow{4}{*}{ViT-Base}  & 8.849 & isoFLOP                       & \textbf{91.57}           &    92.29   & 88.98 \\
                                   & 9.876 & MoD   & 89.8        &  \textbf{92.87}  &  \textbf{92.85} \\
                                   & 9.875 & \ouralgo  & 89.26  & 91.77 &  92.61\\
       \midrule
       \multirow{4}{*}{ViT-Large}  & 33.439 & isoFLOP               &\textbf{92.97}           & \textbf{97.7}     &   \textbf{92.3} \\
                                   & 34.523 & MoD   & 90.87             & 95.85      &  89.1\\
                                   & 34.52 & \ouralgo  & 91.39   & 96.66   & 88.4\\
        \bottomrule
    \end{tabular}
\end{table}

\begin{table}[ht]
    \centering
    \caption{Top-1 accuracy for MoD models with $50\%$ capacity, compared with isoFLOP baselines.}
    \label{tab:accuracy-transfer-50}
    \begin{tabular}{l l c c c c}
        \toprule
        \textbf{Model} & \textbf{FLOPS (G)} & \textbf{Configuration} & \textbf{Cars}  & \textbf{Flowers} & \textbf{Pets}\\
        \midrule
        \multirow{4}{*}{DeiT-Tiny} & 0.951   & isoFLOP                       & \textbf{88.22}            & \textbf{93.42}    &  87.84    \\
                                   & 0.927 & MoD   & 83.98       & 87.82      &  86.94 \\
                                   & 0.927 & \ouralgo  & 86.9   & 89.75 & \textbf{87.92}  \\
        \midrule
        \multirow{4}{*}{DeiT-Small} & 3.47 & isoFLOP                       & \textbf{91.97}          &  \textbf{94.6}    &  	87.84  \\
                                   & 3.42 & MoD   & 89.58               & 91      &  91.36  \\
                                   & 3.42 & \ouralgo  & 90.10   & 91.8 & \textbf{92.12} \\
        \midrule
        \multirow{4}{*}{ViT-Base}  & 13.216 & isoFLOP & \textbf{92.61} & \textbf{96}    &  92.64  \\
                                   & 13.105 & MoD   & 91.18             &  93.62    &  92.72 \\
                                   & 13.104 & \ouralgo  & 91.15   & 93.08 & \textbf{93.21}\\
       \midrule
       \multirow{4}{*}{ViT-Large}  & 46.241 & isoFLOP                       & \textbf{93.17}       & 96.13     &  \textbf{92.8}\\
                                   & 45.927 & MoD   & 92            & 95.69    &  89.67 \\
                                   & 45.925 & \ouralgo  & 91.95  & \textbf{96.56} & 90.37 \\
        \bottomrule
    \end{tabular}
\end{table}

\subsection{Effect of multiplying routing weights}
\begin{figure}[h!]
    \centering
 
    \subfigure[DeiT-T]{ 
        \includegraphics[width=0.3\linewidth]{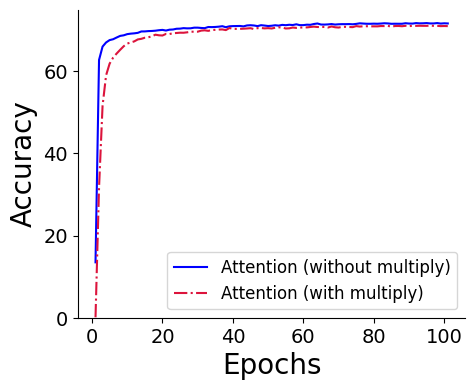}
        \label{fig:convergence-deit-t-multiply}
    }
    \subfigure[DeiT-S]{ 
        \includegraphics[width=0.3\linewidth]{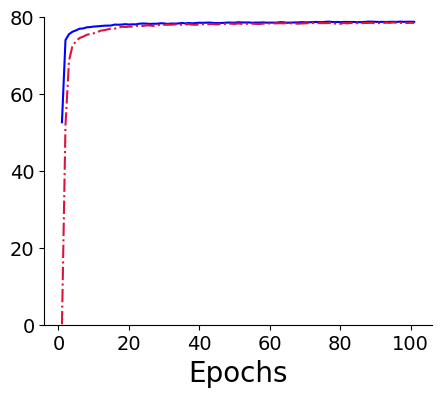}
        \label{fig:convergence-deit-s-multiply}
    }
   \caption{\textbf{\ouralgo with routing scores multiplied to MoD output}. Multiplying the output of the MoD block with routing scores for \ouralgo (red curve) compared to the proposed \ouralgo without multiplication (blue curve).}
    \label{fig:attn-multiply}
\end{figure}

We compare \ouralgo with a modified version which multiplies the attention routing scores to the output of the MoD layer.
Results in \cref{fig:attn-multiply} show that multiplying the routing scores to the output can slow down convergence.

\subsection{Effect of learning rates}

\begin{figure}[ht!]
    \centering
    \subfigure[DeiT-T]{
        \includegraphics[width=0.4\linewidth]{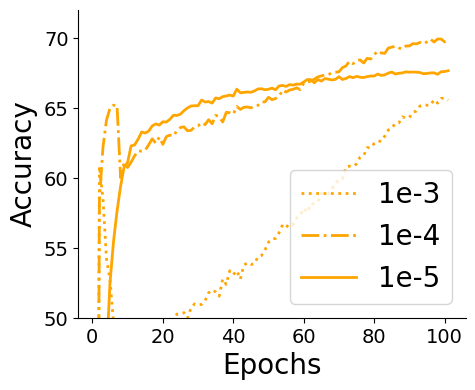}
        \label{fig:lr-deit-t}
    }
    \subfigure[DeiT-S]{
        \includegraphics[width=0.4\linewidth]{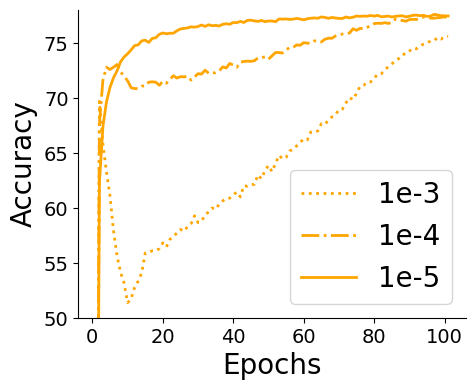}
        \label{fig:lr-deit-s}
    }
    \caption{Sweep over learning rates on ImageNet-1k for standard routing.}
    \label{fig:lr-sweep-deit}
\end{figure}

\begin{figure}[ht!]
    \centering
    \includegraphics[width=0.4\linewidth]{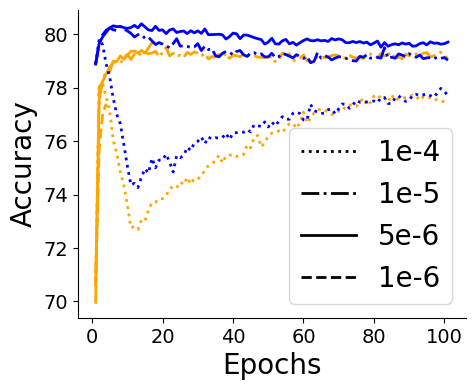}

    \caption{Sweep over learning rates on ImageNet-1k for standard routing and \ouralgo on ViT-Base. Orange curves denote standard routing and blue curves denote attention routing.}
    \label{fig:lr-sweep-vit}
\end{figure}

We identify the optimal learning rates for finetuning by conducting a sweep across a range of learning rates for finetuning on ImageNet-1k as shown in \cref{fig:lr-sweep-deit} and \cref{fig:lr-sweep-vit}.

\subsection{Additional visualizations for \ouralgoheading}
\begin{figure}[ht!]
    \centering
    
    \includegraphics[width=\linewidth]{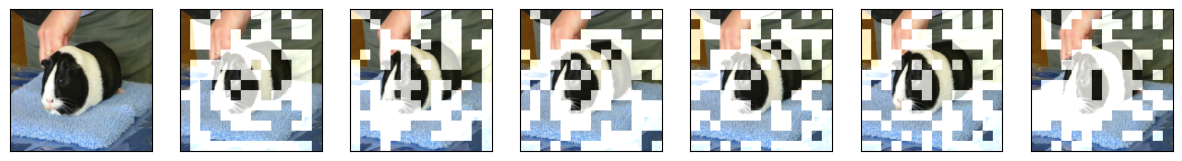}  
    \includegraphics[width=\linewidth]{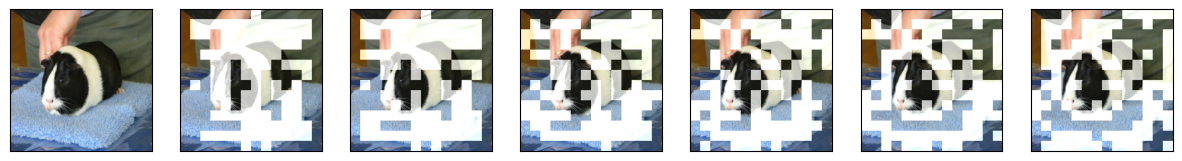}
    \includegraphics[width=\linewidth]{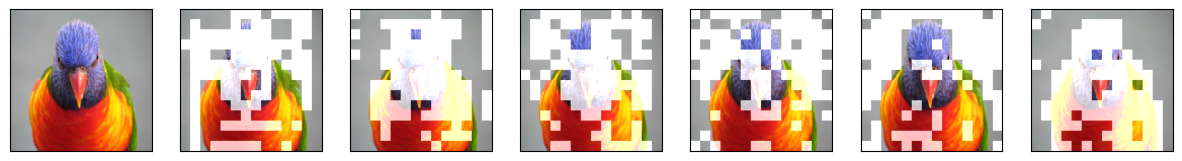}  
    \includegraphics[width=\linewidth]{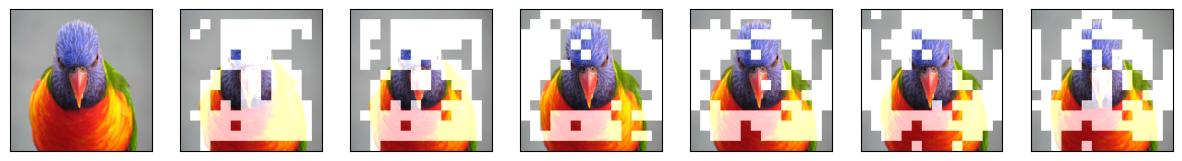}
    
    \caption{Routing example on ViT Base with $50\%$ capacity trained on ImageNet-1k. Each example shows tokens chosen by standard routing (top) and attention routing (bottom). Each column represents a MoD layer as depth increases from left to right.}
    \label{fig:example-routing-vit-b}
\end{figure}

\begin{figure}[ht!]
    \centering
    \includegraphics[width=\linewidth]{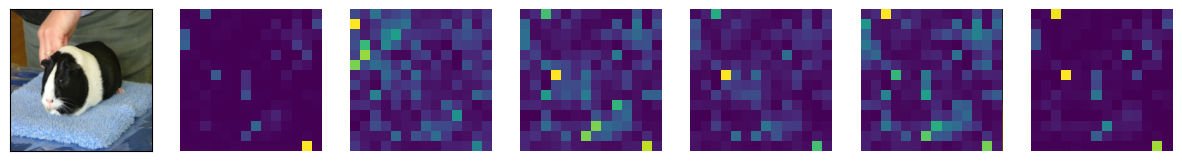}  
    \includegraphics[width=\linewidth]{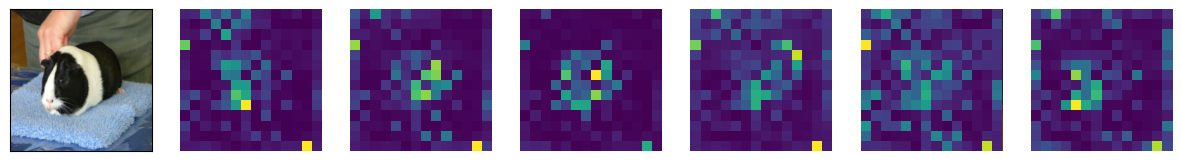} 
    \includegraphics[width=\linewidth]{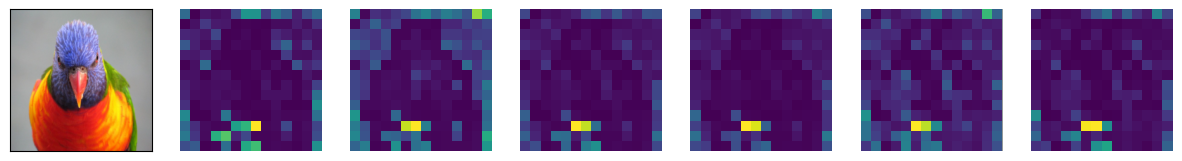}  
    \includegraphics[width=\linewidth]{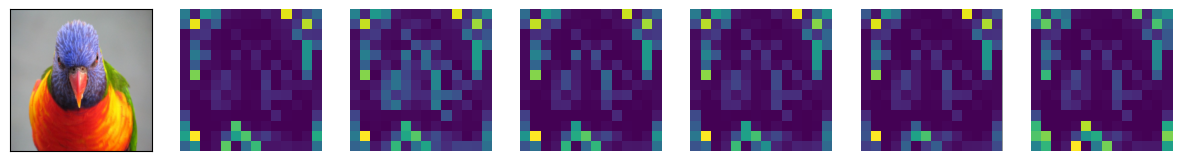} 

    \caption{Attention map of each head in the last layer of a ViT Base MoD with $50\%$ capacity for standard routing (top) and attention routing (bottom) finetuned on ImageNet-1k. Each column denotes an attention head of the last layer.}
    \label{fig:example-attention-vit-b}
\end{figure}

In \cref{fig:example-routing-vit-b} and \cref{fig:example-attention-vit-b} we show the routed patches of each MoD layer and the attention maps of the last MoD layer in a ViT-Base trained on ImageNet respectively.

\begin{figure}[ht!]
    \centering
    \includegraphics[width=\linewidth]{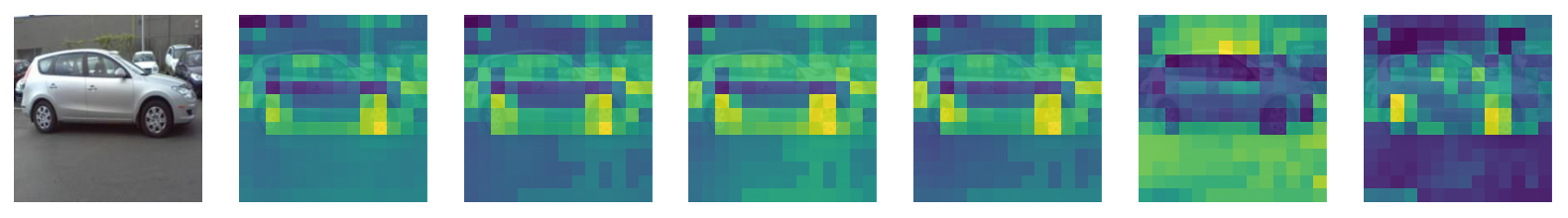}  
    \includegraphics[width=\linewidth]{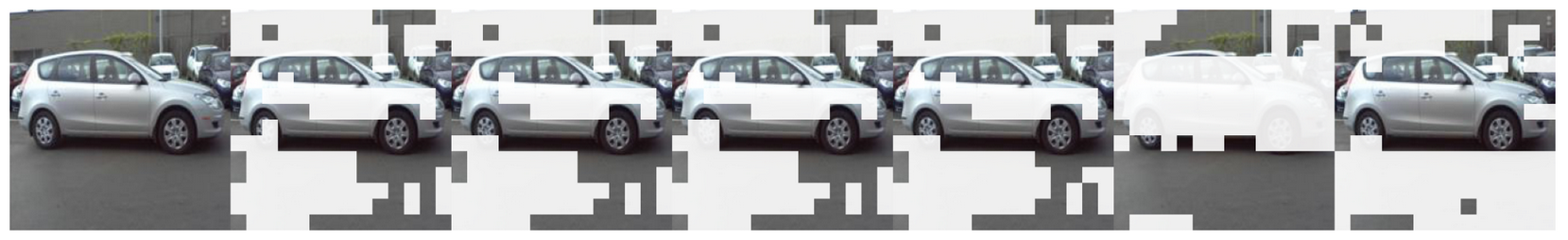} 
    \includegraphics[width=\linewidth]{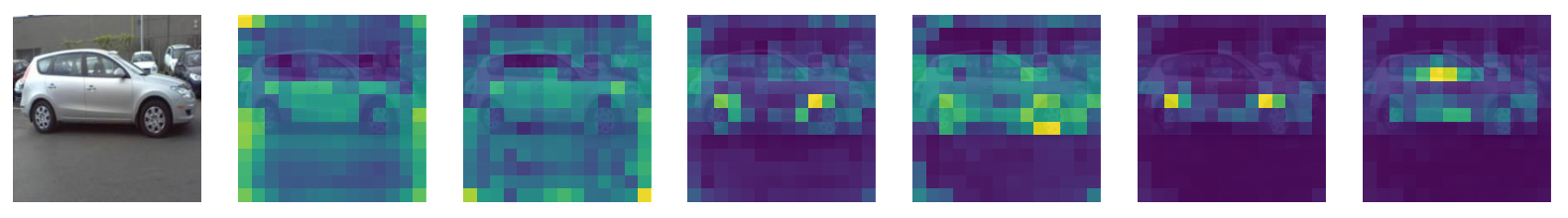}  
    \includegraphics[width=\linewidth]{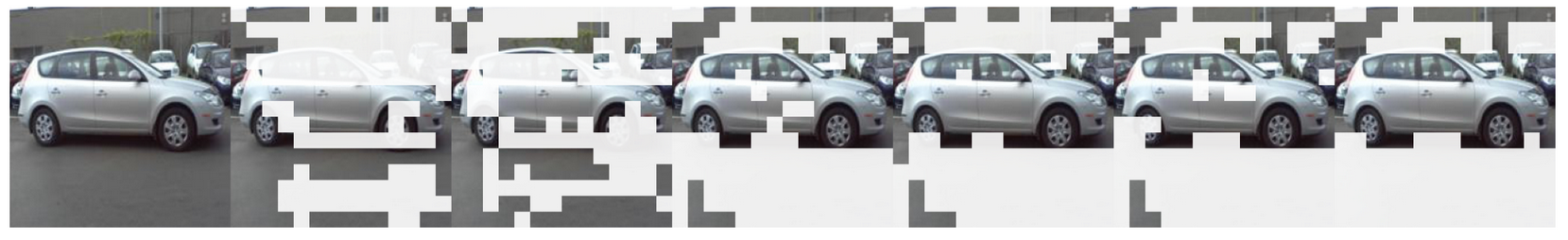} 
    \caption{Visualizing routing scores and token selection of each MoD layer of a DeiT-Tiny MoD with $50\%$ capacity for standard routing (top) and attention routing (bottom) finetuned on Stanford Cars.}
    \label{fig:example-cars-deit-tiny}
\end{figure}

Additionally, we visualize the routed patches and the routing weights for a DeiT-Tiny model trained on Stanford Cars in \cref{fig:example-cars-deit-tiny}.

\subsection{\ouralgo with Flash Attention}

Flash Attention has been proposed by \citet{dao2022flashattention} as a method to implement attention without the need to compute the $N \times N$ attention map explicitly, reducing hardward communication overhead and thus speeding up computation considerably. 
Flash Attention directly computes the final output of attention $A_h^{l}V_h^{l}$ for each head based on intermediate tiling steps which compute the attention map implicitly. We propose an alternate method to perform attention routing with \ouralgo in Flash Attention framework as we do not have access to the attention map in this case. This allows efficient implementation of our method.

Referring to Algorithm 1 in \citet{dao2022flashattention}, our goal is to aggregate the attention scores from each tile $B$ and gather them in a vector of size $ \ermR \in \mathbb{R}^{N \times 1}$ which are the routing weights.
Thus, we can perform attention routing with \ouralgo without explicitly forming the $N \times N$ attention map which can be expensive in terms of computation.

Following similar notations from Algorithm 1 in \citet{dao2022flashattention}, we perform \ouralgo as shown in \cref{alg:flash-attn}. Basically, for each query $\ermQ_i$, $\ermA_{\text{temp}}$ will be used to aggregate nonnormalized attention scores for individual tokens in line 3-7. The scores will then be normalized and accumulate to the final routing weights $\ermR$ in line $8-9$.
It must be noted that we swap the order of tiling from row first (as in \citet{dao2022flashattention}) to column first in order to aggregate row wise scores efficiently. Thus in each iteration only the $Q$ can be cached while the $K$ and $V$ need to be loaded for each tile. Our algorithm introduces a small memory overhead of $O(N)$ due to additional temporary variables.

\begin{algorithm}
\caption{\ouralgo with Flash Attention, modified from Algorithm 1 in \citet{dao2022flashattention}}
\label{alg:flash-attn}
\begin{algorithmic}[1]
\STATE Initialize, $\ermR \in \mathbb{R}^{N \times 1}$ and $\ermA_{\text{temp}} \in \mathbb{R}^{B_r \times N}$ to zero.
\FOR{each $i$ from 1 to $T_r$}
    \FOR{each $j$ from 1 to $T_c$}
        \STATE $\ermS_{ij} = \ermQ_i\ermK_j^T$
        \STATE $\ermP_{ij} = \exp\left(\ermS_{ij} \right)$ 
        \STATE Accumulate for row sum as, $\ermA_{\text{temp},j} = \ermP_{ij}$
    \ENDFOR
    \STATE $\vl_i = \sum_{j} \ermA_{\text{temp},j}, \vl_i \in \mathbb{R}^{Br}$
    \STATE $\ermR \leftarrow \ermR + \ermP_{ij} / \vl_i$ 
\ENDFOR
\STATE \textbf{return} Routing weights $\ermR$
\end{algorithmic}
\end{algorithm}

\subsection{Comparison with token pruning methods}
\begin{figure}
    \centering
    \includegraphics[width=0.75\linewidth]{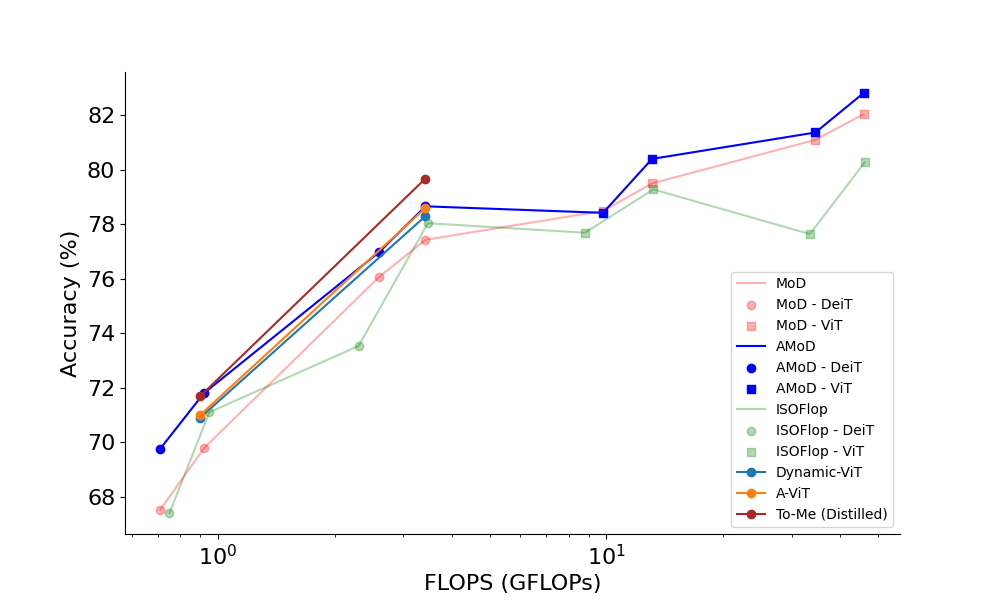}
    \caption{Accuracy Comparison with token-pruning and merging methods}
    \label{fig:comparison_accuracy_token_pruning_methods}
\end{figure}
We compare our method \ouralgo with other token-pruning and token-merging including Token Mergin (ToME) \citep{tome}, A-ViT \citep{yin2022vit} and Dynamic-ViT \citep{rao2021dynamicvit} to validate the performance of \ouralgo.
We compare with the baseline results provided in Table 11 in \citet{tome} and Table 3 in \citet{yin2022vit}.
However, we note that ToMe \citep{tome} trains their models with distillation while the other methods do not, which aids ToMe.
Results are provided in \cref{tab:token-pruning-baseline} and \cref{fig:comparison_accuracy_token_pruning_methods}
\begin{table}[ht!]
    \centering
    \caption{Comparison with other token-pruning and merging methods (* denotes training with distillation).}
    \label{tab:token-pruning-baseline}
    \begin{tabular}{l l c c }
\toprule
\textbf{Model} & \textbf{Method}            & \textbf{Top-1 Acc (\%)} & \textbf{FLOPs (G)} \\ 
\midrule
\textbf{DeiT-T} & A-MoD                      & 71.8                    & 0.9                \\ 
                & A-ViT                      & 71.0                    & 0.8                \\ 
                & Dynamic ViT                & 70.9                    & 0.9                \\ 
                & ToMe (with distillation)   & 71.69*                  & 0.93               \\ 
\midrule
\textbf{DeiT-S} & A-MoD                      & 78.66                   & 3.42               \\ 
                & A-ViT                      & 78.6                    & 3.6                \\ 
                & Dynamic ViT                & 78.3                    & 3.4                \\ 
                & ToMe (with distillation)   & 79.68*                  & 3.43               \\ 
\bottomrule
\end{tabular}
\end{table}

\subsection{Model Throughput}
\begin{figure}
    \centering
    \includegraphics[width=0.75\linewidth]{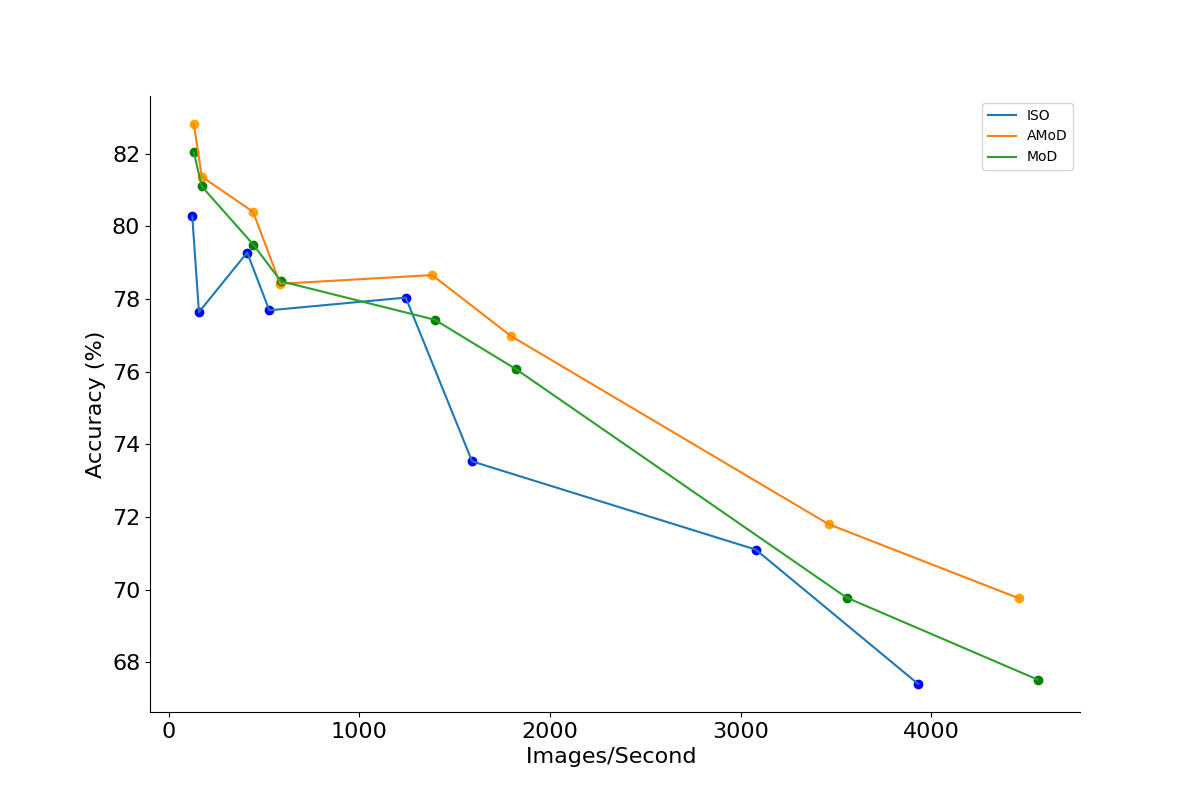}
    \caption{Accuracy vs Throughput for MoD vs ISOFlop Models with Batch Size 100 on Nvidia A100 GPU.}
    \label{fig:comparison_accuracy_throughput}
\end{figure}

\begin{figure}
    \centering
    \includegraphics[width=\linewidth]{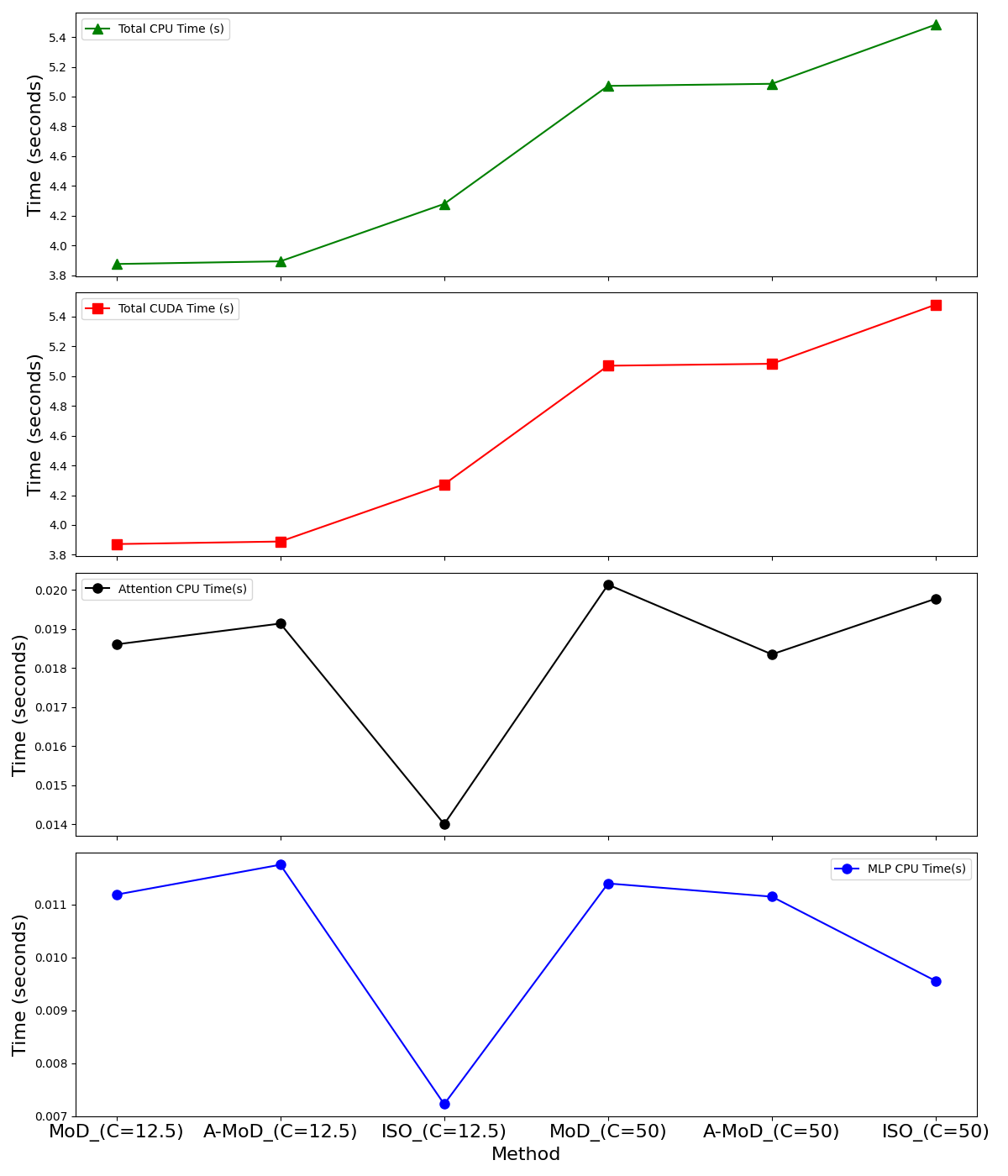}
    \caption{Profiling \ouralgo, MoD and isoFLOP \textbf{ViT-Base} methods on Nvidia A100 GPU. The x-axis shows different models from left to right: MoD, \ouralgo and isoFLOP for both C=$12.5\%$ and C=$50\%$. }
    \label{fig:profiler}
\end{figure}

We provide a comparison of model throughput in \cref{fig:comparison_accuracy_throughput}.
\ouralgo has a higher throughput (img/s) in comparison to MoD and isoFLOP baselines.
We also provide a breakdown of each method using the PyTorch profiler to highlight the CPU and GPU time used by each method for both the Attention layer and the MLP layer as shown in \cref{fig:profiler}.

\subsection{Efficiency of \ouralgo}
To highlight the efficiency of A-MoD, we compare it with the baseline DeiT-S and report the top-1 accuracy on ImageNet. A-MoD is able to reduce the number of FLOPs by up to $18\%$ without dropping performance, with standard training and no additional tricks. Results are provided in \cref{tab:amod-efficiency}.
\begin{table}[h!]
\centering
\caption{Comparison of A-MoD ($70\%$ capacity) with the vanilla DeiT-S baseline which has more FLOPs.}
\begin{tabular}{ l c c }
 \toprule
\textbf{Model}        & \textbf{FLOPs (G)} & \textbf{Top-1 Accuracy (\%)} \\ 
\midrule
DeiT-S Baseline       & 4.6                & 79.6                   \\ 
\midrule
A-MoD (C = $70\%$)                 & 3.8                & 79.63                  \\ 
\bottomrule
\end{tabular}
\label{tab:amod-efficiency}
\end{table}

\subsection{Training from scratch}
We also provide results for training from scratch on ImageNet and observe that A-MoD outperforms standard routing as shown in \cref{tab:train-from-scratch}.

\begin{table}[h!]
\centering
\caption{Training from scratch comparison for \ouralgo and MoD on ImageNet-1k.}
\begin{tabular}{l c c c}
\toprule
\textbf{Model}            & \textbf{Training Epochs} & \textbf{Method} & \textbf{Accuracy (\%)} \\ 
\midrule
\multirow{2}{*}{DeiT-S}                   & \multirow{2}{*}{300}             & A-MoD           & 76.63                 \\ 
                         &                 & MoD             & 75.90                 \\ 
\midrule
\multirow{2}{*}{ViT-Base}                 & \multirow{2}{*}{160}             & A-MoD           & 73.66                 \\ 
                         &                 & MoD             & 72.47                 \\ 
\bottomrule
\end{tabular}
\label{tab:train-from-scratch}
\end{table}

\subsection{Results on object detection}
We also provide results with the DETR architecture \citep{detr} for \ouralgo.
\cref{tab:detr_comparison} shows that MoD and \ouralgo achieve comparable results in this case.

\begin{table}[h!]
\centering
\caption{Comparison of mAP and FLOPs across DETR MOD, DETR \ouralgo, and DETR-Baseline.}
\begin{tabular}{l c c c}
\toprule
\textbf{Model}                              & \textbf{MoD, C=50\%} & \textbf{\ouralgo, C=50\%} & \textbf{Baseline} \\ 
\midrule
\textbf{mAP}                                 & 39.6\%               & 38.6\%                 & 39.9\%                 \\ 

\textbf{FLOPs(G) (Total)}                      & 83.2                 & 83.2                   & 86.56                  \\ 

\textbf{FLOPs(G) (Transformer Encoder/Decoder)} & 7.747                & 7.745                  & 10.745                 \\ 
\bottomrule
\end{tabular}
\label{tab:detr_comparison}
\end{table}

\end{document}